\documentclass[preprint,12pt,3p,twocolumn]{elsarticle}

\usepackage[english]{babel}
\usepackage{graphicx}%
\usepackage{pst-all, graphics, graphicx, color}
\usepackage{subcaption}
\usepackage{amsmath, amssymb} 
\usepackage{epstopdf}
\usepackage{algorithm, algpseudocode}
\usepackage{booktabs}
\usepackage{multirow}
\usepackage[nolist]{acronym}

\newacro{dmp}[DMP]{Dynamic Movement Primitive}
\newacro{seds}[SEDS]{Stable Estimator of Dynamical Systems}
\newacro{gmm}[GMM]{Gaussian Mixture Model}
\newacro{gmr}[GMR]{Gaussian Mixture Regression}
\newacro{gp}[GP]{Gaussian Process}
\newacro{tpgmm}[TP-GMM]{Task-Parameterized \ac{gmm}}
\newacro{lfd}[LfD]{Learning from Demonstration}
\newacro{il}[IL]{Imitation Learning}
\newacro{wls}[WLS]{weighted least-squares}
\newacro{vs}[VS]{Visual Servoing}
\newacro{ds}[DS]{Dynamical System}
\newacro{nn}[NN]{Neural Network}
\newacro{eqln}[EQLN]{Equation Learner Network}
\newacro{tpeqln}[TP-EQLN]{Task-Parameterized Equation Learner Network}
\newacro{mp}[MP]{Movement Primitive}
\newacro{cnmp}[CNMP]{Conditional Neural Movement Primitive}

\usepackage{todonotes}

\definecolor{darkgreen}{rgb}{0.0,0.49,0.19}

\usepackage{multicol}

\journal{Robotics and Autonomous Systems}

\renewcommand{\vec}[1]{\boldsymbol{#1}}

\begin{document}
\begin{frontmatter}


\title{Learning and Extrapolation of Robotic Skills\\ using Task-Parameterized Equation Learner Networks}
\author[1]{Hector Villeda\corref{cor1}}
\ead{Hector.Villeda@uibk.ac.at}
\author[1,2]{Justus Piater} 
\ead{Justus.Piater@uibk.ac.at}
\author[3]{Matteo Saveriano}
\ead{Matteo.Saveriano@unitn.it}

\cortext[cor1]{Corresponding author}
\address[1]{Department of Computer Science, University of Innsbruck, Innsbruck, Austria}
\address[2]{Digital Science Center (DiSC), University of Innsbruck, Innsbruck, Austria.}
\address[3]{Department of Industrial Engineering, University of Trento, Trento, Italy}

\begin{abstract}
%
Imitation learning approaches achieve good generalization within the range of the training data, but tend to generate unpredictable motions when querying outside this range. We present a novel approach to imitation learning with enhanced extrapolation capabilities that exploits the so-called \ac{eqln}. Unlike conventional approaches, \acp{eqln} use supervised learning to fit a set of analytical expressions that allows them to extrapolate beyond the range of the training data. We augment the task demonstrations with a set of task-dependent parameters representing spatial properties of each motion and use them to train the \ac{eqln}. At run time, the features are used to query the \ac{tpeqln} and generate the corresponding robot trajectory. The set of features encodes kinematic constraints of the task such as desired height or a final point to reach. 
We validate the results of our approach on manipulation tasks where it is important to preserve the shape of the motion in the extrapolation domain. Our approach is also compared with existing state-of-the-art approaches, in simulation and in real setups.
The experimental results show that \ac{tpeqln} can respect the constraints of the trajectory encoded in the feature parameters, even in the extrapolation domain, while preserving the overall shape of the trajectory provided in the demonstrations. 
\end{abstract}




\end{frontmatter}




\section{Introduction}\label{sec:intro}

\ac{il}, also known as  robot programming by demonstration, is a powerful approach that allows the robot to learn new skills from human guidance~\cite{billard2016learning}. The main advantage of this approach is that the user can instruct a robot with no knowledge about programming. New tasks are learned using only a set of demonstrations that the user gives as input to the learning algorithm. Task demonstrations can be provided, for instance,  using kinesthetic teaching~\cite{saveriano2015incremental, caccavale2019kinesthetic} where the user physically guides the robot towards the task execution. 

\begin{figure}[t]
    \centering
    \includegraphics[width=\columnwidth]{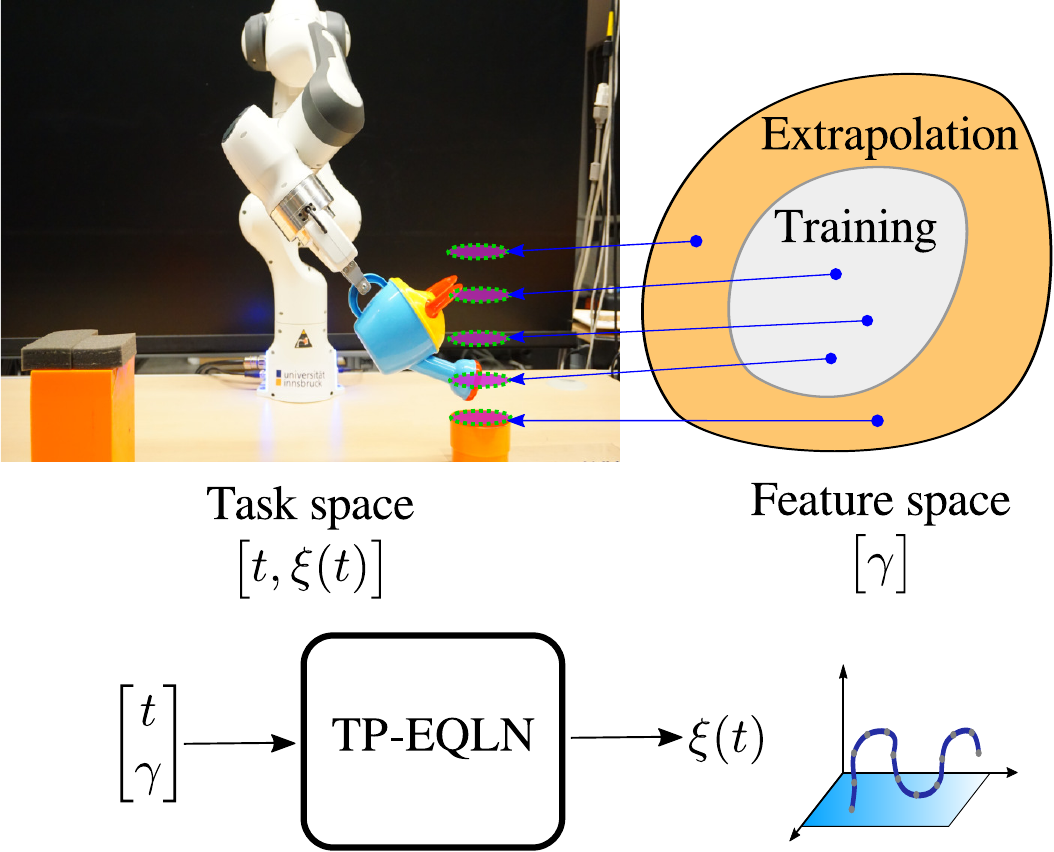}
    \caption{\textit{Pouring task}. In this example, each parameter value $\gamma$ corresponds to a different size of the cup, and $\xi(t)$ corresponds to the parameterized trajectory at time $t$ that the robot has to perform in order to pour the water. }
    \label{fig:TP_EQLN_Operation}
\end{figure}

Task demonstrations represent successful executions of a given task in a certain range of operative conditions. However, everyday tasks often demand good generalization capabilities in order to successfully execute the same task in different conditions. As an example, consider the pouring task in  Fig.~\ref{fig:TP_EQLN_Operation}. In this case, the user can reasonably demonstrate how to pour into a few cups of different size and, from this data, the robot should infer how to pour into unobserved cups of different sizes. Task generalization can be achieved by enriching the demonstrated motion trajectories with further task-dependent features that characterize the task execution~\cite{ude2010task, calinon2016tutorial, pervez2018learning}. Task-dependent features, also called  \textit{task parameters}, are given as further input to regress a feasible motion for the given executive context (e.g., a cup of different size). Task-parameterized approaches for \ac{il}, as other data-driven approaches, produce effective motions inside the range of the demonstrations. However, robotic tasks often demand good generalization capabilities outside the range of training data.

In this paper, we present a novel approach to generating task-parameterized motions that uses a special type of neural networks called \acf{eqln}~\cite{MartiusL16, pmlr-v80-sahoo18a}. \acp{eqln} combine a set of different basic functions to fit the training data with analytical expressions. This is a key difference with standard learning approaches that use one type of activation function like Gaussian, sigmoid, or ReLu. As a result, the expressions fit by the \ac{eqln} have better generalization capabilities and extrapolate well beyond the range of the training data. Following the idea of task parameters, we enrich the demonstrations with extra features describing spacial properties of the motion like the size of a box or a certain height to reach. This leads to the \ac{tpeqln} approach proposed in this work.

\acp{tpeqln} cope with two main problems related to task-parameterized approaches: 1) They achieve good extrapolation performance over the space of task parameters while 2) preserving the overall shape of the demonstrations that defines the task by itself. In several robotic tasks, it is crucial to consider these mentioned problems in order to ensure a successful execution. We compare the performance of \acp{tpeqln} with existing approaches and further demonstrate the effectiveness of our approach with physics-based simulations and in a real setup on a set of manipulation tasks.

Section~\ref{sec:rel_work} reviews related work. Section.~\ref{sec:method} presents the proposed approach for motion extrapolation. In Sec.~\ref{sec:experiments}, we present an experimental evaluation and a comparison with existing approaches. 
Section~\ref{sec:conclusion} states the conclusions and proposes further extensions.

\section{Related Work}\label{sec:rel_work}
The \ac{il} formalism is built around the concept of \ac{mp} intended as a compact and flexible representation of the observed motion which allows generalization to different scenarios and reduces memory requirements~\cite{Calinon2007learning}. A variety of approaches exist to represent \acp{mp} and each of them has distinctive features and limitations. We provide a review of existing approaches that specifically focus on task generalization. 

Stable \acp{ds} are capable of planning converging motions from any point of the state space and have been effectively used as \ac{mp}~\cite{Khansari2011learning, Perrin16fast, saveriano2018incremental, saveriano2020energy}. Among the \ac{ds}-based representations, \acp{dmp}~\cite{Ijspeert2013Dynamical, Saveriano2021Dynamic} are certainly the most popular and several authors have investigated their generalization capability. A \ac{dmp} encodes a demonstrated trajectory into a spring-damper dynamics with a nonlinear forcing term. The forcing term acts as a disturbance on the linear dynamics and allows the \ac{dmp} to generate arbitrarily complex trajectories. A set of learnable shape parameters are used to parameterize the forcing term. The shape parameters determine the overall shape of the \ac{dmp} trajectories and play an important role in the generalization to new situations. Ude et al.~\cite{ude2010task} use a \ac{gp}~\cite{Rasmussen2006} to learn and retrieve the mapping between a query point (typically a different goal) and the \ac{dmp} shape parameters. Instead of learning a mapping that outputs a new set of \ac{dmp} parameters, the approach in~\cite{Stulp2013learning} embeds the task parameters directly into the forcing term. Pervez et al.~\cite{pervez2018learning} follow a similar idea, but use a \ac{gmm}~\cite{cohn1996active} to represent the forcing term. Their experiments shows better generalization performance compared to the approaches in~\cite{ude2010task, Stulp2013learning}. 
Alternative approaches provide a probabilistic characterization of the learned \ac{mp}~\cite{calinon2016tutorial, paraschos2013probabilistic, huang2019kernelized, Huang2020Toward, Zhou2019Learning}. Probabilistic \acp{mp}~\cite{paraschos2013probabilistic} perform generalization by statistical conditioning on the query task parameters, typically a new goal or a set of via-points. Via-point \ac{mp}~\cite{Zhou2019Learning} and Kernelized \ac{mp}~\cite{huang2019kernelized, Huang2020Toward} have better extrapolation capabilities than Probabilistic \ac{mp}, but also in these approaches task parameters are new goals to reach or a set of via-points to traverse. The \ac{tpgmm}~\cite{calinon2016tutorial} considers as task parameters the homogeneous transformations between arbitrary reference frames. By observing the human demonstrations from each of these frames the robot is able to learn the spatial relationship between start, goal, and via-points in the trajectory.  


Going beyond the kinematic level, Abu-Dakka et al.~\cite{abu2015adaptation} use \acp{dmp} to regulate the contact force during an interaction task, while Yang et al.~\cite{Yang2018dmp} exploit \acp{dmp} to generalize a set of learned, variable-impedance robotic skills. Teaching variable impedance and force skills is also a challenge, due to the couplings between robot, environment, and human teacher~\cite{abu2020variable}. Rozo et al.~\cite{Rozo2016Learning} exploit teleoperation and a simplified model of the environment to extract force profiles and build training data. Kronander and Billard~\cite{kronander2013learning} use an artificial skin to wiggle the robot and reduce its stiffness. Electromyography (EMG) is another widely-used human-robot teaching interface, with applications in prosthetic hand interaction control~\cite{jaquier2021tensor, fang2022semg}, co-manipulation~\cite{peternel2018robot}, and impedance learning~\cite{Zeng2021Simultaneously}.

Conditional Neural Processes~\cite{garnelo2018conditional} are a class of deep \acp{nn} that combines the function approximation power of \ac{nn} with the data efficiency of Bayesian approaches like \ac{gp}. Inspired by conditional neural processes, Seker et al.~\cite{seker2019conditional} developed an imitation learning framework called \ac{cnmp}. \ac{cnmp} generates motion trajectories by sampling observations from the training data and predicting a conditional distribution over target  points. Training data may include robot position, forces, and any task parameters. \ac{cnmp} is promising but have limited extrapolation capabilities. A possibility to improve the extrapolation performance is to combine imitation and reinforcement learning~\cite{akbulut2020acnmp}. However, we seek a solution within the imitation learning framework.     


Standard regression techniques fit the training data by combining a single type of activation functions through learnable weights. For example \ac{gmm} uses Gaussians while \ac{nn} uses sigmoid or ReLu functions. We claim that the poor extrapolation performance of standard regression approaches arises form this design choice. \acfp{eqln}~\cite{MartiusL16, pmlr-v80-sahoo18a} are designed to combine a set of activation functions in order to fit the training data with a richer analytical representation that possibly matches the underlying function we are training to approximate from observations. In other words, an \ac{eqln} attempts to find equations from observations. Symbolic regression approaches attempt to find such equations by searching in a certain function space by means of evolutionary algorithms~\cite{schmidt2009distilling}. This is a key difference between symbolic regression  and \ac{eqln}, which uses standard backpropagation techniques. In this work, we extend the \ac{eqln} and present a novel \ac{il} approach called \acf{tpeqln}.


\section{Methods}\label{sec:method}
\subsection{Problem definition}

In Task-parameterized \acp{mp}, the task to be learned can be defined as a set of tuples $\Phi=\{\vec{\gamma}_i,\xi(t)_i\}_i^{M}$, for $i = 1,\ldots, M$ and $t=0,\ldots,T_i$, where $M$ is the number of demonstrated trajectories $\xi_i(t)\in\mathbb{R}^{D}$ and $T_i$ is the time duration of each trajectory. For simplicity, we assume that the demonstrations have the same length, i.e., that each $T_i = T$. For each demonstrated trajectory $\xi_i(t)$, there exists a set of $n$ feature parameters $\vec{\gamma}_i\in\mathbb{R}^{n}$ associated with each $\xi_i(t)$. Task parameters encode information of the spatial behavior and the task's shape, e.g., a certain height to reach.  As an example, in the pouring task shown in Fig. \ref{fig:TP_EQLN_Operation} the cup's size (task parameter) determines the arc length, its initial and final point as well as the orientation trajectory that the end effector of the robot has to perform in order to pour into the cup successfully. Therefore, task-parameterized \acp{mp} try to learn a mapping between the parameter space and the trajectory space.


%
\begin{figure*}[t]
    \centering
    \includegraphics[width=0.85\textwidth]{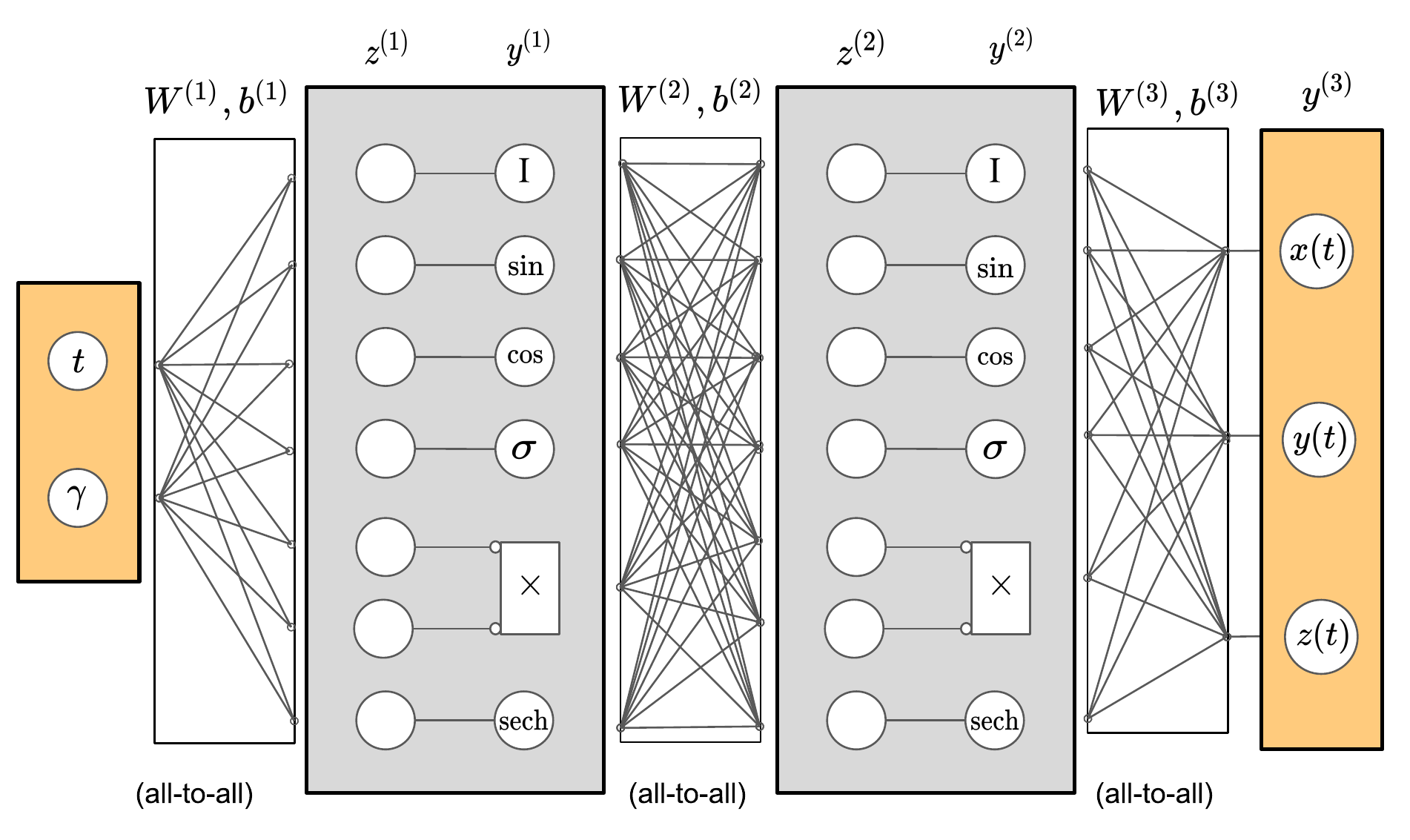}
    \caption{\ac{tpeqln} architecture: The network receives as run-time input the time step and the values of the feature parameters. The network outputs the pose of the end effector at each time step.} 
    \label{fig:TP-EQLN_architecture}
\end{figure*}
Let us to define $\zeta$ as the whole parameter space for a given task $\Phi$. In imitation learning, the user can only demonstrate a subset of the whole parameter space  Fig.~\ref{fig:TP_EQLN_Operation}. 
When a query parameter $\vec{\gamma}$ is sampled from $\zeta$, we interpolate to regress the corresponding trajectory $\xi(t)$. Traditional regression approaches perform well in interpolation. However, in real situations, we are interested in querying outside the demonstration area. In this case, we need to extrapolate to regress the corresponding trajectory. This problem is hard to solve due to the lack of training data in the extrapolation domain. The goal of this paper is to learn a mapping between task parameters and motion trajectories that achieves good inter- and extrapolation performance. As mentioned in Sec.~\ref{sec:intro}, we build our \ac{tpeqln} on the \ac{eqln} approach originally proposed by~\cite{MartiusL16}.

\subsection{TP-EQLN architecture}\label{sec:FNN_Architecture}
An \ac{eqln} is a multi-layered feed-forward network with $L$ layers; $l=\left \{1, ..., L-1\right \}$  are hidden layers, and the last layer $L$ is an output layer. The input  $z^{l}\in\mathbb{R}^{u}$ to the $l$-th layer is defined as
\begin{equation}\label{eq:z}
\begin{split}
    z^{(l)} &= W^{(l)}y^{(l-1)}+b^{(l)} \\
    &= W^{(l)}f_a\left({z^{(l-1)}}\right)+b^{(l)},
\end{split}
\end{equation}
where $y^{(l-1)} \in \mathbb{R}^{v} $ is the output of the preceding layer $l-1$, $W^{(l)}\in\mathbb{R}^{u\times v}$ is a matrix of weights, and $b^{(l)}\in\mathbb{R}^{u}$ is a bias vector. $y^{(0)}$ and $y^{(L)}$ are defined as the network input and output respectively. $W$ and $b$ are the open parameters to learn. A traditional \ac{nn} uses a single kind of activation function $f_a$, while the \ac{eqln} uses the following activation functions: 
\begin{equation}
\begin{array}{rclcl}
f_0(z)&=& I(z) &=& z\\ 
f_1(z)&=&\sin(z)\\
f_2(z)&=&\cos(z)\\
f_3(z)&=&\sigma(z) &=& \frac{1}{1+\text{e}^{-z}}\\
f_4(z_{0},z_{1})&=&z_{0} \times z_{1}\\ 
f_5(z)&=& \text{sech}(z) &=& \frac{2}{\text{e}^{z}+\text{e}^{-z}}\\
\end{array}
\label{eq:activation_functions}
\end{equation}
For a multidimensional input the above activation functions are applied element-wise. Note that the original formulation~\cite{MartiusL16} uses only a subset of Eq.~\eqref{eq:activation_functions}, namely $f_0$ through $f_4$. The original layers are redefined by including new activation functions that contribute to better performance, as experimentally shown in the ablation study in Section~\ref{subsec:ablation}.
For the output $y^{(L)}$, the same configuration is kept as defined in~\cite{MartiusL16}, a fully-connected layer with linear activation functions. In contrast, the first layer $y^{(0)}$ is modified to include time and feature parameter(s), i.e. the time vector $\vec{t}=0,\ldots,1$ is concatenated to the task parameter $\vec{\gamma}$ to form the input $\mathcal{I}=\left[ \vec{\gamma},\vec{t}\right ]$ of the \ac{tpeqln}, where $\mathcal{I}\in\mathbb{R}^{n+1}$.
Therefore, the output of the network $y^{(L)}$ represents the estimated trajectory $\hat{\xi}(t)$ at each time step $t$ for a queried feature parameter $\mathcal{I}$.
The \ac{tpeqln} architecture used in this work is shown in Fig.~\ref{fig:TP-EQLN_architecture}.

\subsection{Cost Function}\label{subsec:cost function}
For training the \ac{tpeqln}, we use the cost function
\begin{multline}\label{eq:loss_function}
    L=\frac{1}{N}\sum_{i=1}^{\left | D \right |} \left \|  \xi_i(t)-\hat{\xi}_i(t) \right \|^{2}+\\ \lambda_{W,b} \sum_{l=1}^{L} \left ( \left | W^{(l)} \right|_1+\left | b^{(l)} \right|_1 \right )+\lambda_P P_\delta
\end{multline}
where the second term is the conventional regularizer that minimizes the weights in order to achieve good generalization. $\lambda_{W,b} \geq0$ defines a regularization constant. We additionally introduce a Penalty Epoch Term 
\begin{multline}\label{eq:Penalty_term}
   P_{\delta}=\sum_{i=1}^{N}\sum_{i=i}^{D} \text{max}\left(\delta_{\xi_{min}}-y_{i}^{L}(\mathcal{I}_{i}), 0\right)\\+\text{max}\left(y_{i}^{L}(I_{i})-\delta_{\xi_{max}}, 0\right).
\end{multline}
to penalize every $k$ epochs the predictions $\hat{\xi}_i(t)$ that are beyond a predefined threshold $\delta_\xi$. $\lambda_P >  0$ defines a regularization constant for the penalty term. The introduction of a priory information in the penalty term helps to restrict the convergence space of the model by penalizing predictions $\hat{\xi}(t)$ that are beyond the boundaries defined by $[\delta_{\xi_{min}},\delta_{\xi_{max}}]$. This leads to better prediction in the extrapolation domain. For robotic manipulation tasks, the restricted space $[\delta_{\xi_{min}},\delta_{\xi_{max}}]$ is given by the workspace of the robot.
To calculate the penalty epoch we randomly sample $N$ inputs $\mathit{I}$ without labels $\xi(t)$. The values for $t$ are sampled from the predefined range $[0,T]$ whereas the values for $\vec{\gamma}$ are sampled considering only the extrapolation domain of the parameter space $\zeta$. 

Regarding the rest of the parameters in the cost function, the boundaries of the parameter space $\zeta$ depends on each task. We use $\lambda_{W,b}=4e^{-3}$, $\lambda_{P}=1e^{-6}$, $k=50$, and $[\delta_{\xi_{min}},\delta_{\xi_{max}}]$ are platform/hardware dependent and are defined as $\delta_{\xi_{min}}=\left [-0.9,-0.9,-0.1 \right ]^{T}$,  and $\delta_{\xi_{max}}=\left [0.9,0.9,1.2 \right ]^{T}$.

The parameters $\theta$ of the network are updated by stochastic gradient descent with mini-batches and the Adam optimizer. We use a learning rate of $\alpha=4e^{-3}$ for all the experiments.
\subsection{Training Phases}\label{subsec:training Phases}

We define a total of $E$ epochs for training the model and split them into 3~phases:

\paragraph{Phase 1: No regularization} This phase takes place within the epochs $0\leq E_i\leq\frac{1}{4}E$: and the cost function to be minimized is $L$ without regularization, i.e., $\lambda_{W,b}=0$. The reason of considering $\lambda_{W,b}=0$ only for a few epochs at the beginning of the training is to allow the weights to freely converge towards a first approximation.
    
\paragraph{Phase 2: Lasso-Like}  This phase takes place within the epochs $\frac{1}{4}E< E_i\leq\frac{3}{4}E$, and the cost function includes the regularization term, with $\lambda_{W,b}=4e^{-3}$.
    The goal of this phase is to enhance the generalization and prevent overfitting.
    
\paragraph{Phase 3 Weight pruning Weights}  This phase performs magnitude-based weight pruning and gradually zeroes out model weights by following the update rule
    \begin{multline}\label{eq:3er_training_Phase}
    W_{i}^{(l)},b_{i}^{(l)}:=\\
    \left\{\begin{array}{cl}
    0& \textup{\;if\;} \left |W_{i}^{(l)}\right|, \left | b_{i}^{(l)} \right | <\delta_{W},
    \\ 
    W_{i}^{(l)},b_{i}^{(l)}& \textup{\; otherwise.}
    \end{array}\right.
    \end{multline}
    This phase takes place within the range $\frac{3}{4}E<E_i\leq T$ and leads to shorter analytic expressions that still properly describe the training data. We use $\delta_{W}=0.01$

The epochs range $E$ used in each training phase was chosen based on the values proposed in \cite{MartiusL16}. We slightly modified the last range to $\frac{3}{4}E$ 
with the aim of increasing sparsity.

A summary of the overall training process is presented in Algorithm~\ref{alg:TP-EQLN_training}.
\begin{algorithm}
\caption{TP-EQLN training}\label{alg:TP-EQLN_training}
\begin{algorithmic}[1] 
\State Take a set of demonstrations $D=\{\vec{\gamma}_i,\xi_i(t)\}$
\State Generate the data set $\mathcal{I}=[\vec{\gamma},\vec{t}]$ for the input and $\xi(t)$ for the target values.
\State Train the \ac{tpeqln} using the loss function from  Eqn. \eqref{eq:loss_function}
\While{$T <  E$}
\If{$T<\frac{1}{4}E$ }
    \State Set $\lambda_{W,b}=0$
\ElsIf{$T<\frac{15}{20}E$}
    \State Set $\lambda_{W,b}\neq0$
\Else
    \State Set $\lambda_{W,b}=0$ and include the update rule from  Eqn. \eqref{eq:3er_training_Phase}
\EndIf
\EndWhile
\end{algorithmic}
\end{algorithm}

\subsection{Extrapolation example}
\begin{figure}[t]
    \centering
    \includegraphics[width=1.0\columnwidth]{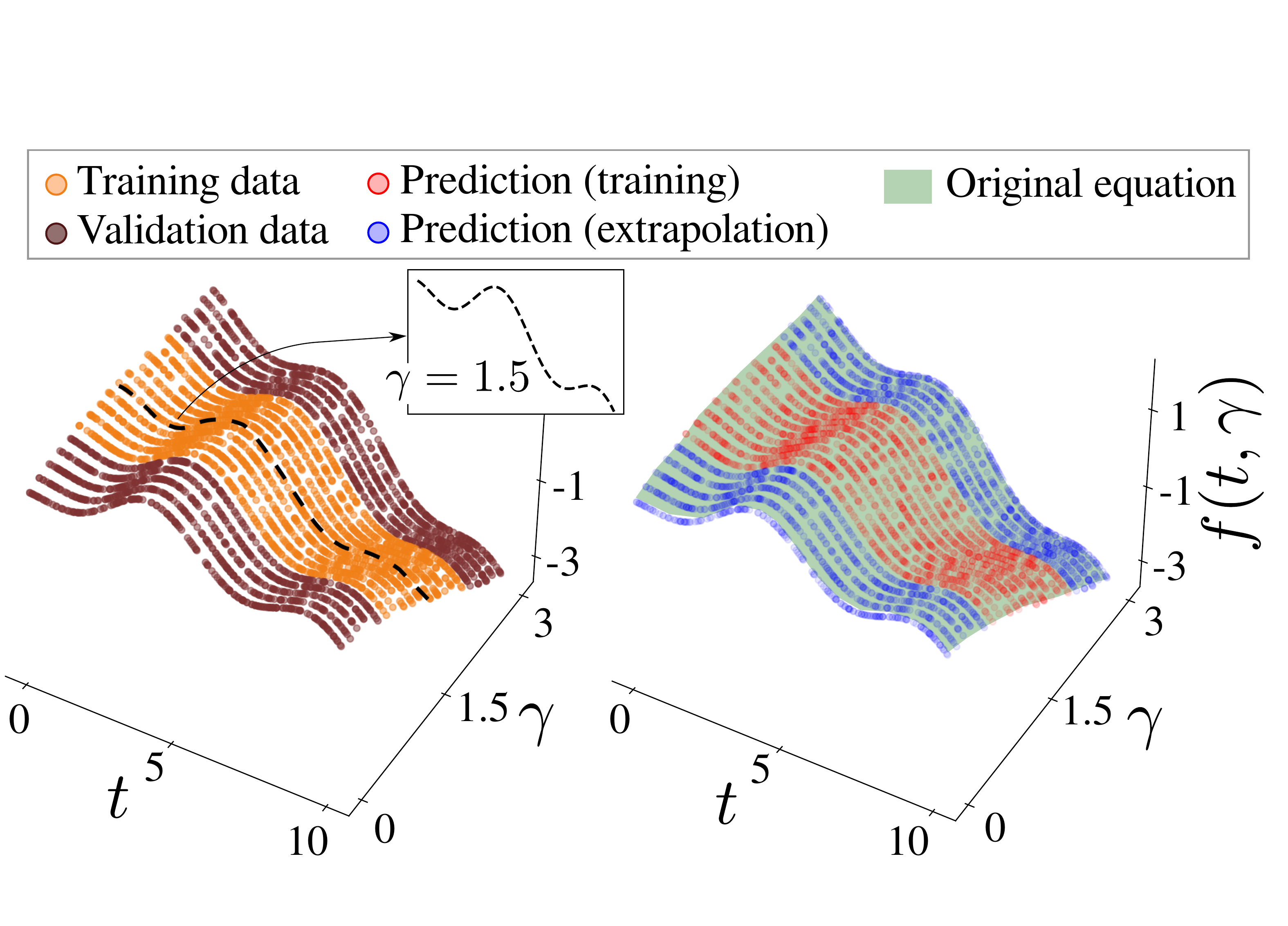}
    \caption{A 1-D toy example that shows the extrapolation capabilities of the \ac{tpeqln}. (Left) the training and extrapolation data sets taken from the original equation. (Right) the prediction of the \ac{tpeqln} for each data set.} 
    \label{fig:Toy_example}
\end{figure}

In this subsection we give an overview of the \ac{tpeqln} performance with a $1D$ toy example. The aim is to find the set of equations via \ac{tpeqln} that better explains the training data set, where the samples are generated as
\begin{equation}\label{eq:eq_to_learn}
\begin{split}
f(t,\gamma)=&-0.024t^2-0.064\gamma^2+0.064 t\\
&-0.112\gamma t+0.256\gamma\\
&-0.5\cos(1.2t-0.15\gamma-1.8)\\
&-0.2\sin(-1.4\gamma-0.8)\\
&+1.3\sigma(1.4t+0.5\gamma)
\end{split}
\end{equation}
Here $f:\mathbb{R}^{2}\rightarrow \mathbb{R}$, $t$ is the time variable and $\gamma$ the feature parameter. The samples  are generated within the range $0\leq \gamma\leq 3$ with 20 equally-spaced subintervals. For each value of $\gamma$, we generate $200$ samples from $f(t,\gamma)$ with $t \sim  U(0,10)$. This gives a total of $4000$ samples.  To test the extrapolation performance, we train the model only considering $10$ samples of $\gamma$ generated within the range $0.789\leq \gamma\leq 2.210$. The remaining samples are used to validate the prediction of \ac{tpeqln} in the extrapolation domain. The training and validation data set is presented in Fig.~\ref{fig:Toy_example}~(left).

We use one hidden layer with a fully-connected layer in the output, with a batch size of $150$. The equation obtained from \ac{tpeqln} after the training process is
\begin{equation}\label{eq:learned_eq}
\begin{split}
f(t,\gamma)=&-0.023t^{2}-0.091\gamma^2+0.064t\\
&-0.111\gamma t+0.543\gamma-0.349\\
&-0.455\cos(-1.048\gamma-0.064)\\
&-0.5\sin(-1.2t+0.15\gamma-3.37)\\
&+1.299\sigma(1.402t+0.5\gamma).
\end{split}
\end{equation}
Comparing the estimated and the ground-truth equation~(\ref{eq:eq_to_learn}), we observe that both are composed of the same trigonometric functions, and $f_5(z)$ does not play any role in either equation. The main differences of the constants are in the  $\sin$ and $\cos$ functions, which could be simplified using the trigonometric identities  $\sin(\theta)=\cos(\theta-\frac{\pi}{2})$ and $\cos(\theta)=\sin(\theta+\frac{\pi}{2})$, in order to obtain an expression more similar to reference equation. Small inaccuracies in the estimated constant factors give rise to mild deviations in the values of the function values that increase as the values of the arguments $t$ and $\gamma$ move away from the trained ranges. However, for the purpose addressed in this work, the error that may be produced is acceptable since the exploration area of interest is limited to the robot's workspace, which remains constant.  The evaluation of the equation estimated by \ac{tpeqln} with training and validation data sets is presented in Fig.~\ref{fig:Toy_example}~(right) indicated by the red and blue data points respectively. The light-green surface represents the real equation from \ref{eq:eq_to_learn}. Here, we observe that the evaluation of both training data sets remains close to the surface of the ground-truth equation. 

\subsection{Ablation study}\label{subsec:ablation}
\begin{figure}[t]
    \centering
    \includegraphics[width=1.0\columnwidth]{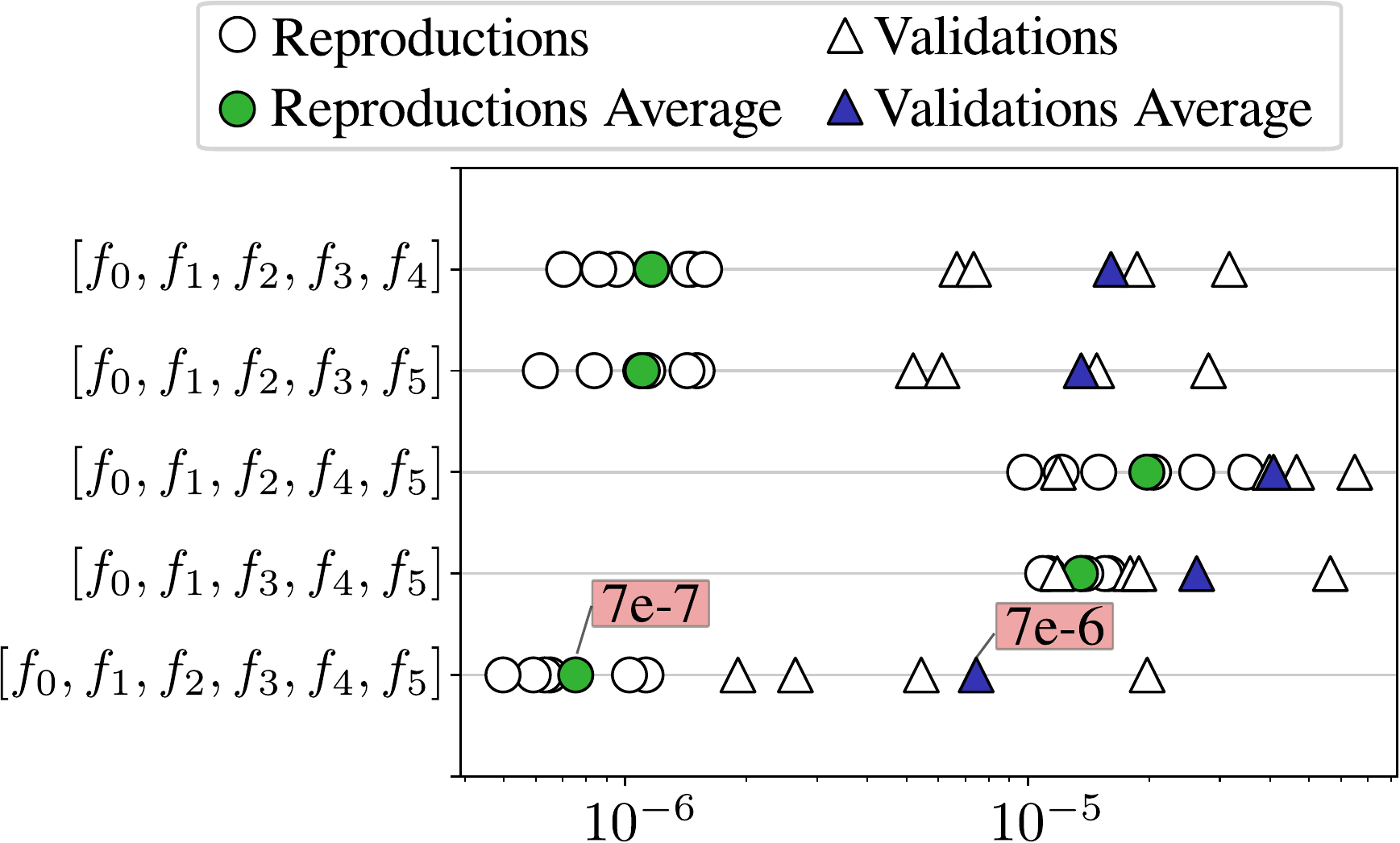}
    \caption{Results of the ablation study.} 
    \label{fig:ablation_study}
\end{figure}

To justify the use of the set of activation functions $[f_0,f_1,f_2,f_3,f_4,f_5]$ in~\eqref{eq:activation_functions} proposed for \ac{tpeqln}, we carried out an ablation study where the performance is analyzed when some key activation functions are removed such as $[f_2,f_3,f_4,f_5]$.  The study is carried out using the data set from task $\Phi_1$. The details of this task are presented in Sec.~\ref{Task_1}. Figure~\ref{fig:ablation_study} presents the results for each removed function. Each of the markers represents the MSE between the demonstrated trajectories and the predictions of each method. The circular markers refer to the training data set, whereas triangular markers refer to validation data set. The green markers represent the MSE of the demonstrations from the training data set and the blue markers represent the MSE for the extrapolation data set. 

The lowest MSE obtained from this study for both the training and the extrapolation data set is highlighted in the light-red boxes, which corresponds to the use of the entire set of activation functions. Using the whole set of activation functions proposed in Section \ref{sec:FNN_Architecture} gives the best results.

\begin{figure}[t]
    \centering
    \includegraphics[width=1.0\columnwidth]{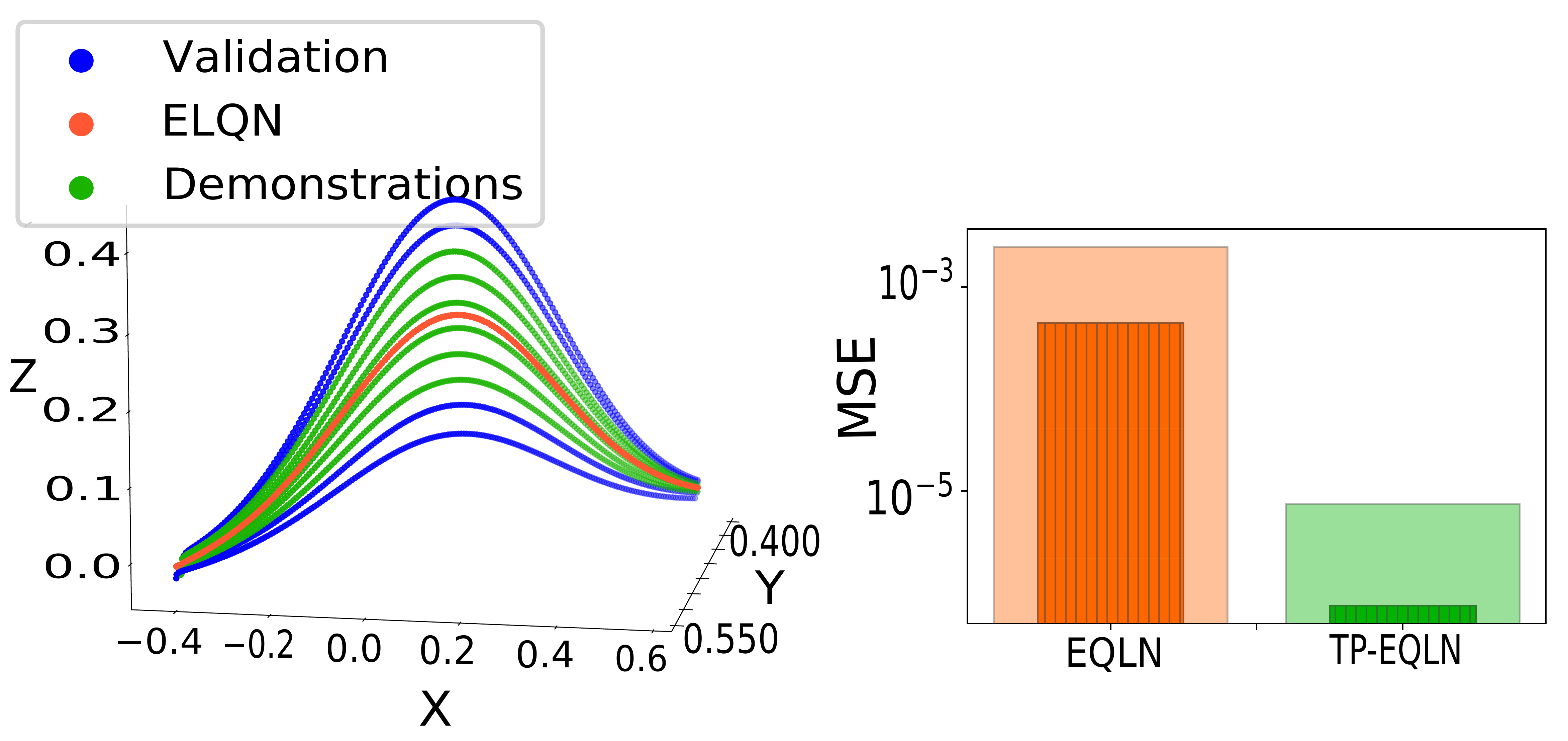}
    \caption{Comparison between \ac{eqln} and \ac{tpeqln}. (Left) The training and validation data set and the predicted trajectory from \ac{eqln}. (Right) The MSE comparison obtained from \ac{eqln} and \ac{tpeqln} respectively. Solid bars correspond to the MSE from the interpolation data set whereas light bars correspond to the MSE from the extrapolation data set.}
    \label{Behavior_cloning}
\end{figure}

We also compared \ac{tpeqln} against the standard \ac{eqln} to show the advantage of adding a feature parameter to the input. As for the ablation study, we used the obstacle avoidance task presented in Sec.~\ref{Task_1}. We used the same architecture for both \ac{tpeqln} and \ac{eqln}, without considering the feature parameter in \ac{eqln}. The results are shown in Fig.~\ref{Behavior_cloning}. Figure~\ref{Behavior_cloning}~(left) shows the trajectories used for training (green) and validation (blue). The orange trajectory is the output of the \ac{eqln}. Since it takes only the time as input, the \ac{eqln} is not able to generalize over different obstacle heights and all trajectories collapse into one. The MSE values for the training and extrapolation data set of each approach are shown in Fig.~\ref{Behavior_cloning} (right), which validate the improvement of the \ac{tpeqln} over the \ac{eqln}.
In conclusion, it is essential to include a feature parameter in the \ac{eqln} to encode and generalize over any physical and or spatial properties of the task. 

\section{Experimental Results}\label{sec:experiments}
We perform six sets of experiments in different scenarios. The first four were carried out in simulation and the last two on a real Franka Emika Panda robot. In each experiment, we evaluate the inter- and extrapolation performance of the proposed \ac{tpeqln}. Our approach is compared against two prominent approaches, namely \ac{cnmp} and \ac{tpgmm}. The estimated trajectory in all the experiments is defined in the Cartesian space. 
First, a brief introduction to the six experiments is given. Afterwards, each experiment is discussed in detail in their respective subsection.

The first three experiments resemble a pick and place task where the robot picks an object from point A and places it at point B. For each experiment different task parameters are used. For the first experiment $\Phi_{1}$, we varied only the obstacle height. In the second experiment $\Phi_{2}$, we varied the height of the obstacle and the endpoint B. In the third experiment $\Phi_{3}$, we vary the height and position of the obstacle and the endpoint B. Demonstrations for $\Phi_1$ to $\Phi_3$ are synthetic data sampled from the parametric equation:
\begin{equation}\label{eq:eq_update_exp}
\begin{bmatrix}
x_r(t)\\
y_r(t)\\
z_r(t) 
\end{bmatrix}=
\begin{bmatrix}
x_0+t\cos{\theta}\\
y_0+t\sin{\theta}\\
z_0+z_d(t) 
\end{bmatrix},
\end{equation}
where the initial position is $\left [x_0,y_0,z_0\right ]^{\top}=\left [0.57,-0.41,0.0\right ]^{\top}$m, $\theta=100\,$deg is the rotation between the trajectory and the global frame, and $t$ and $z_d(t)$ are vectors that parameterize the motion of each task; $t$ varies linearly ($200$ samples) in the range reported in Table~\ref{table:Feature_parameters_tasks}, while $z_d(t)$ is sampled from the Gaussian distribution $z_d(t) = \gamma_1 \left(2\pi\sigma^2\right)^{-\frac{1}{2}}\exp\left(-\frac{(t-\mu)^2}{2\sigma^2}\right)$. The parameter $\gamma_1$ varies the height of the Gaussian. The ranges of $\gamma_1$, $\mu$ and $\sigma$ are listed in Table~\ref{table:Feature_parameters_tasks}. 

The fourth experiment $\Phi_{4}$ consists of opening boxes of different sizes by holding their cover and executing  an  arch  trajectory  until the  box  is entirely open. The fifth experiment $\Phi_{5}$ is carried out in a real setup with a Panda robot and consists of picking and placing an object into different containers while avoiding some fixed obstacles in the environment. 
The last experiment $\Phi_{6}$ is also carried out in a real setup and consists of pouring water into cups of varying size.

\ac{tpeqln} hyperparameters used in each experiments are shown in Table~\ref{table:Hyperparameteres_OB_1D}. The values were selected empirically using the following rationale. Selecting a small number of hidden layers leads to better generalization in extrapolation but loses details in fitting the trajectory. On the other hand, selecting a large number of hidden layers fits the training data better but leads to highly nonlinear curves in the extrapolation domain, which impedes generalization in the extrapolation in some cases. As a general rule, one can select a high number of hidden layers if the trajectories in the extrapolation domain are expected to  be highly nonlinear. Otherwise, a low number of hidden layers is preferable. We found that $1$ to $3$ hidden layers are enough to capture the trajectories of the performed experiments. The batch sizes are normally low for simple trajectories that do not demand large amounts of training data. The number of epochs is directly related to the size of the data set and the number of feature parameters. We found that after $20.000$ epochs the prediction accuracy does not change significantly. A small number of epochs could be enough for simple trajectories that present low nonlinear variations between the feature parameters as in the case of $\Phi_4$.

For \ac{tpgmm} we used $30$ Gaussian components. This number was chosen empirically by varying the number of components between $3$ and $40$ and checking the reproduction accuracy. After $30$ components the estimated trajectory showed some overfitting. Moreover, we use $3$ reference frames (beginning, middle, and end of the trajectory) for $\Phi_1$ to $\Phi_3$, and  $2$ reference frames (beginning and end of the trajectory) for $\Phi_5$ and $\Phi_6$. 
For \ac{cnmp} we used $4$ layers with $128$ units per layer for both the encoder and decoder parts of the network, $3\times10^6$ training steps, and an Adam optimizer with $l_r=10^{-4}$. For \ac{tpeqln} and \ac{cnmp} the same input data format $\mathcal{I}$ was used for the training in all the experiments.
\begin{table*}[t]
 \caption{Feature parameters used for each task.}
 \centering
 \label{table:Feature_parameters_tasks}
 \resizebox{\textwidth}{!}{
 \setlength{\tabcolsep}{0.8\tabcolsep}
\begin{tabular}{|c|c|c|c|c|c|c|c|c|} 
\hline
\multirow{2}{*}{\textbf{Task}} & \multirow{2}{*}{\begin{tabular}[c]{@{}c@{}}\textbf{Feature} \\\textbf{Parameter}\end{tabular}} & \multirow{2}{*}{} & \multirow{2}{*}{\textbf{Feature Space}}   & \multicolumn{2}{c|}{\textbf{Data set}} & \multicolumn{3}{c|}{\textbf{Motion parameters}}  \\ \cline{5-9}  &  &  & & \textbf{Training}  & \textbf{Extrapolation} &$\mu$&$\sigma$&t   \\ 
\hline
\multirow{2}{*}{$\Phi_1$} & \multirow{2}{*}{$\vec{\gamma}\in\mathbb{R}$}  & \textbf{Ranges}   & $0.085\leq\gamma_1\leq0.4$ &$0.155\leq \gamma_1\leq 0.33$    & Remaining &\multirow{2}{*}{$0.5$}&\multirow{2}{*}{$0.21$}&\multirow{2}{*}{$0\leq t\leq 1$}      \\ 
\cline{3-6}
\multicolumn{1}{|l|}{}    &  & \textbf{samples}  & 10   & 6   & 4 & & &  \\ 
\hline

\multirow{2}{*}{$\Phi_2$} & \multirow{2}{*}{$\vec{\gamma}\in\mathbb{R}^{2}$}   & \textbf{Ranges}   & \begin{tabular}[c]{@{}c@{}}$0.085\leq \gamma_1\leq 0.26$,\\$-0.6\leq\gamma_2\leq-0.3$\end{tabular} & \begin{tabular}[c]{@{}c@{}}$0.12\leq\gamma_{1}\leq0.225$,\\$-0.54\leq\gamma_{2}\leq-0.36$\end{tabular} & Remaining  &\multirow{2}{*}{$0.5$}&\multirow{2}{*}{$0.12$}&\multirow{2}{*}{$0\leq t\leq 0.5-\gamma_2$}      \\
\cline{3-6}   &     & \textbf{samples}  & 36 & 16  & 20 & & &    \\
\hline

\multirow{2}{*}{$\Phi_3$}                       & \multirow{2}{*}{$\vec{\gamma}\in\mathbb{R}^{3}$}  & \textbf{Ranges}   & \begin{tabular}[c]{@{}c@{}}$0.085\leq\gamma_1\leq0.365$,\\$-0.6\leq\gamma_2\leq-0.35$,\\ $-0.2\leq\gamma_3\leq0$\end{tabular} & \begin{tabular}[c]{@{}c@{}}$0.15\leq\gamma_1\leq0.29$,\\ $-0.53\leq\gamma_2\leq-0.41$,\\ $-0.15\leq\gamma_3\leq-0.05$\end{tabular} & Remaining &\multirow{2}{*}{$0.5+\gamma_3$}&\multirow{2}{*}{$0.12$}&\multirow{2}{*}{$0\leq t\leq 0.5-\gamma_2$}      \\
\cline{3-6}   &     & \textbf{samples}  & 125 & 98  & 27 & & &    \\
\hline

\multirow{2}{*}{$\Phi_4$}                       & \multirow{2}{*}{$\vec{\gamma}\in\mathbb{R}$}  & \textbf{Ranges}   & \begin{tabular}[c]{@{}c@{}}$0.06\leq \gamma\leq 0.6$ \end{tabular} & \begin{tabular}[c]{@{}c@{}}$0.24\leq \gamma\leq 0.48$\end{tabular} & Remaining &\multirow{2}{*}{--}&\multirow{2}{*}{--}&\multirow{2}{*}{--}      \\
\cline{3-6}   &     & \textbf{samples}  & 10 & 5  & 5  & & &  \\
\hline

\multirow{2}{*}{$\Phi_5$}                       & \multirow{2}{*}{$\vec{\gamma}\in\mathbb{R}^{6}$}  & \textbf{Ranges}   & \begin{tabular}[c]{@{}c@{}}$\gamma_1=0.37$,\\$\gamma_2=-0.52$,\\$\gamma_3=0.12$\\$0.28\leq\gamma_4\leq0.59$,\\ $0.0\leq\gamma_5\leq52$,\\ $0.15\leq\gamma_6\leq0.19$\end{tabular} & \begin{tabular}[c]{@{}c@{}}$C_1=[\gamma_1,\gamma_2,\gamma_3, \gamma_4=0.56$,\\$\gamma_5=0.0$,$\gamma_6=0.17]$\\$C_2=[\gamma_1,\gamma_2,\gamma_3, \gamma_4=0.56$,\\$\gamma_5=0.17$,$\gamma_6=0.17]$\\ $C_3=[\gamma_1,\gamma_2,\gamma_3, \gamma_4=0.29$,\\$\gamma_5=0.03$,$\gamma_6=0.17]$\\ $C_4=[\gamma_1,\gamma_2,\gamma_3, \gamma_4=0.29$,\\$\gamma_5=0.21$,$\gamma_6=0.17]$\end{tabular} & \begin{tabular}[c]{@{}c@{}}$C_5=[\gamma_1,\gamma_2,\gamma_3, \gamma_4=0.59$,\\$\gamma_5=0.32$,$\gamma_6=0.19]$\\$C_6=[\gamma_1,\gamma_2,\gamma_3, \gamma_4=0.58$,\\$\gamma_5=0.50$,$\gamma_6=0.15]$\\ $C_7=[\gamma_1,\gamma_2,\gamma_3, \gamma_4=0.28$,\\$\gamma_5=0.35$,$\gamma_6=0.17]$\\ $C_8=[\gamma_1,\gamma_2,\gamma_3, \gamma_4=0.29$,\\$\gamma_5=0.52$,$\gamma_6=0.16]$\end{tabular}  &\multirow{2}{*}{--}&\multirow{2}{*}{--}  &\multirow{2}{*}{--}    \\ 
\cline{3-6}   &     & \textbf{samples}  & 8 & 4  & 4 & & &   \\
\hline

\multirow{2}{*}{$\Phi_6$}                       & \multirow{2}{*}{$\vec{\gamma}\in\mathbb{R}$}  & \textbf{Ranges}   & \begin{tabular}[c]{@{}c@{}}$0.075\leq \gamma\leq 0.245$ \end{tabular} & \begin{tabular}[c]{@{}c@{}}$0.12\leq \gamma\leq 0.19$\end{tabular} & Remaining &\multirow{2}{*}{--}&\multirow{2}{*}{--}  &\multirow{2}{*}{--}    \\  
\cline{3-6}   &     & \textbf{samples}  & 5 & 3  & 2 & & &     \\
\hline
\end{tabular}
}
\end{table*}
\begin{table}[t]
\centering
 \caption{\ac{tpeqln} hyperparamers used for each task.}
\label{table:Hyperparameteres_OB_1D}
\resizebox{\columnwidth}{!}{  
 \begin{tabular}{cccc} 
 \toprule
 Task & Layers / Batch / Epochs & \multicolumn{1}{c}{Functions per layer}& \\
 \midrule
$\Phi_1$  & $3$ / 150 / 20\,000 & $\phantom{1\times{}}\{I, \sin, \cos, \sigma, \times, \text{sech}\}$\\ 
$\Phi_2$  & $3$ / 150 / 20\,000 & $\phantom{1\times{}}\{I, \sin, \cos, \sigma,  \times, \text{sech}\}$\\
$\Phi_3$  & $3$ / 150 / 20\,000 & $\phantom{1\times{}}\{I, \sin, \cos, \sigma, \times, \text{sech}\}$\\
$\Phi_4$  & $1$ / \phantom{0}50 / \phantom{0}5\,000 & $2\times\{I, \sin, \cos, \sigma,  \times, \text{sech}\}$\\
$\Phi_5$  & $2$ / 150 / 20\,000 & $2\times\{I, \sin, \cos, \sigma,  \times, \text{sech}\}$\\
$\Phi_6$  & $1$ / 100 / 20\,000 & $3\times\{I, \sin, \cos, \sigma,  \times, \text{sech}\}$\\
\bottomrule
\end{tabular}
}
\end{table}

For tasks $\Phi_1$ to $\Phi_5$, the orientation of the end-effector was not considered since it was not relevant for the tasks. Therefore, the estimated trajectory for these tasks is  $\hat{\xi}(t)\in\mathbb{R}^{3}$ (end-effector position). The orientation ($4$D unit quaternion) is considered in task $\Phi_6$, resulting in an estimated trajectory  $\hat{\xi}(t)\in\mathbb{R}^{7}$. The feature parameters and the training and extrapolation data set ranges of the experiments are summarized in Table~\ref{table:Feature_parameters_tasks}.

\subsection{$\Phi_1$ - Variable height} \label{Task_1}
\begin{figure}[t]
    \centering
    \includegraphics[width=0.48\columnwidth]{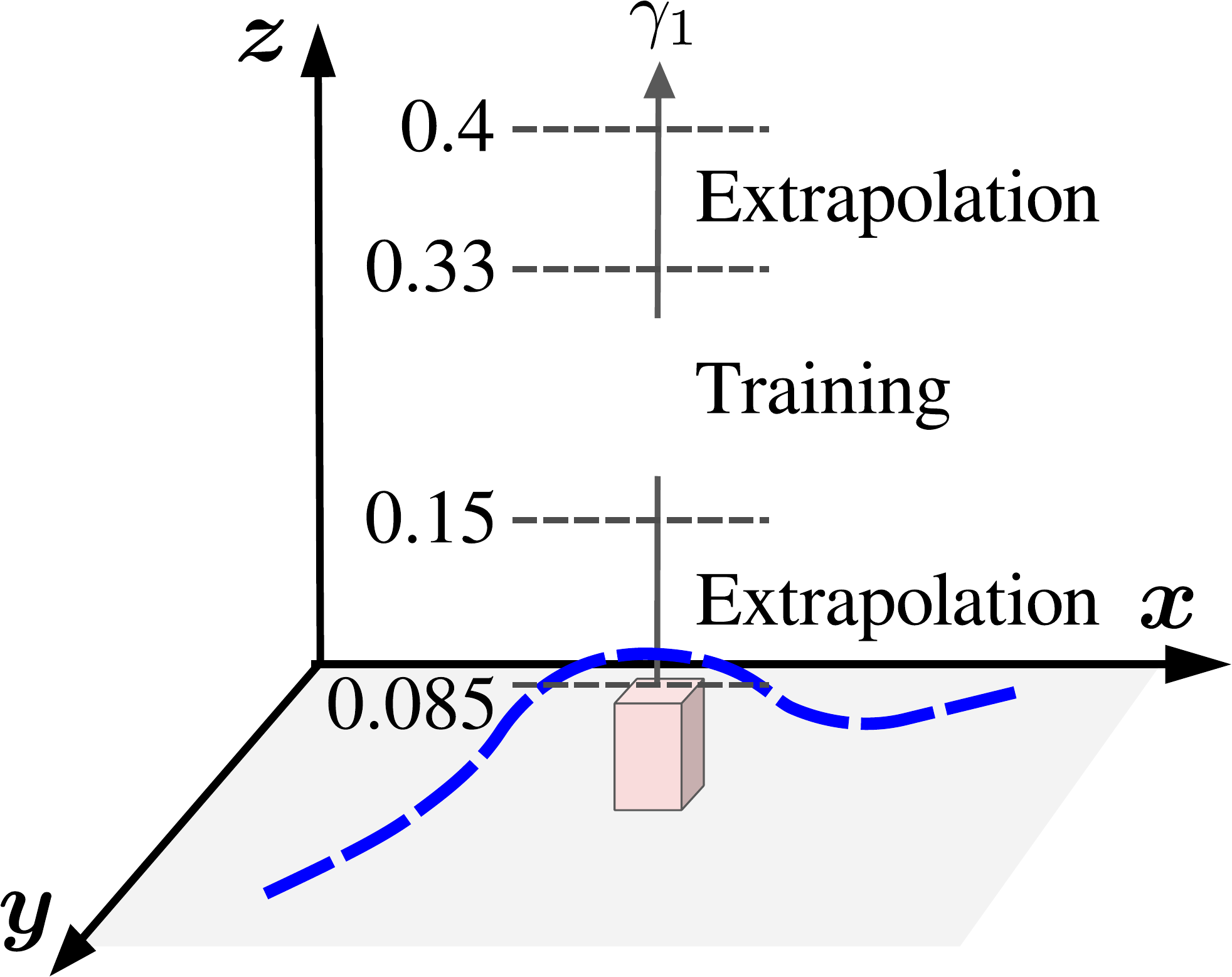}
    \includegraphics[width=0.48\columnwidth]{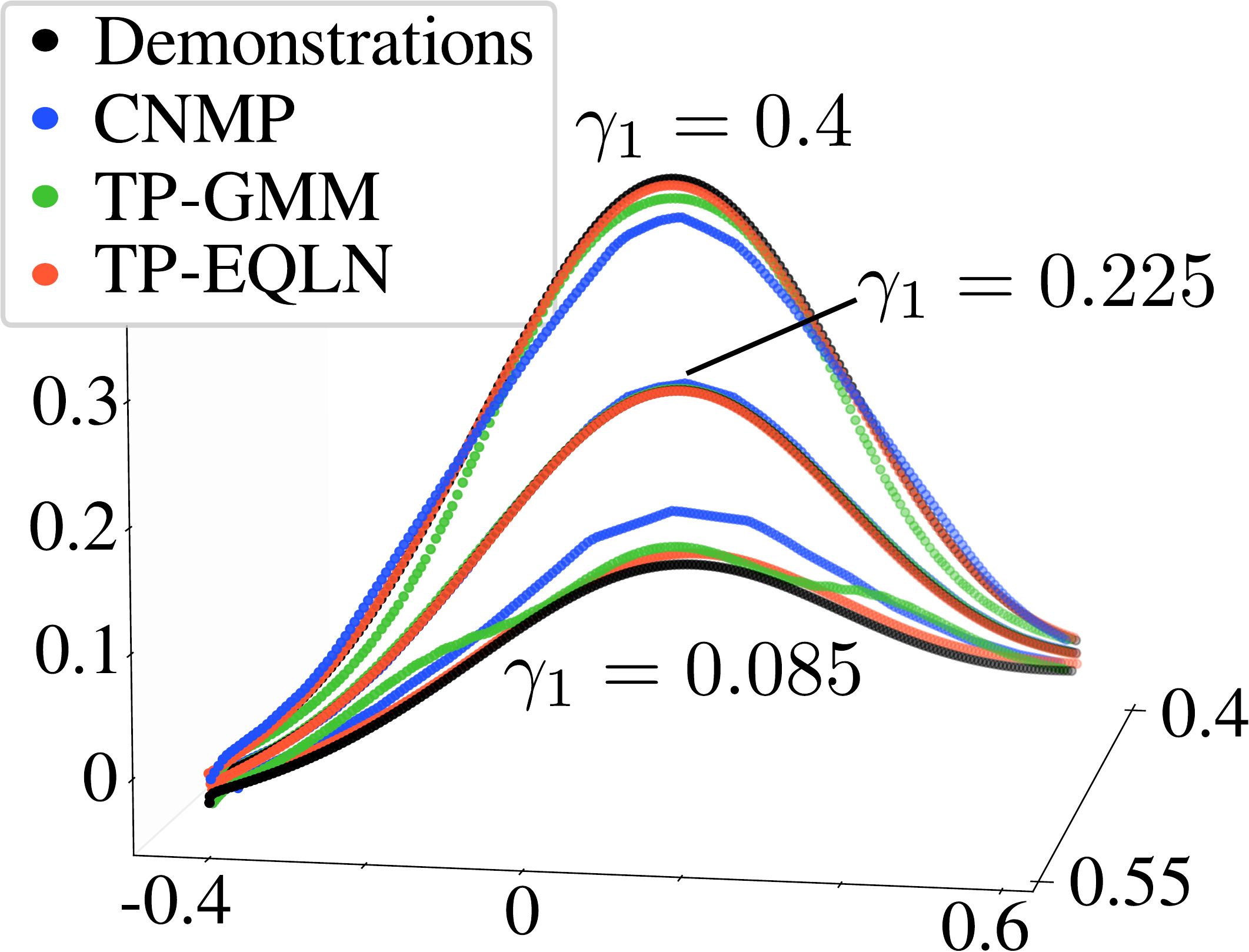}
    
    \vspace{3mm}
    
    \includegraphics[width=1.0\columnwidth]{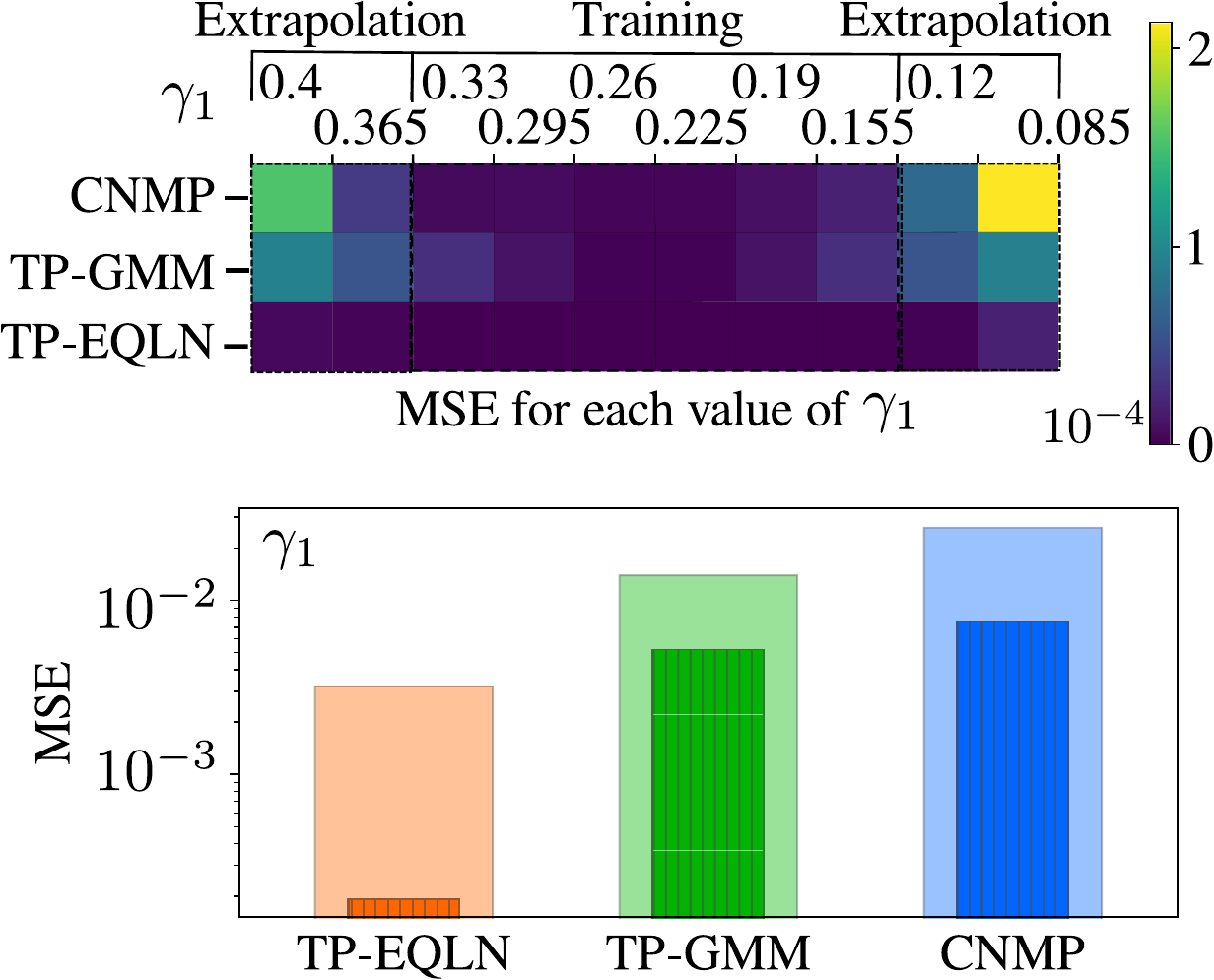}    
    \caption{Definition of the task $\Phi_1$ and qualitative results. (Top-left) The feature space considered for the task. (Top-right) Three trajectories, one from the training set and two for the validation set, obtained for each method. (Bottom) MSE obtained for each method considering all the demonstrations in the $\Phi_1$ task. Shaded (non-shaded) areas indicate the extrapolation (training) domain.}
    \label{fig:results_OB_1D}
\end{figure}

This task consists of avoiding an obstacle of varying height $h$, with only one feature parameter $\gamma_1 = h$, Fig.~\ref{fig:results_OB_1D}~(top-left). The range of $\gamma_1$ and the training and extrapolation data sets are reported in  Table~\ref{table:Feature_parameters_tasks}. 
The position of the obstacle is fixed at $[0.5, 0.07, 0.0]\,$m for all the demonstrations. 


We show three different trajectories for each approach in Fig.~\ref{fig:results_OB_1D}~(top-right), two for $\gamma_1=[0.08,0.4]$ (extrapolation set) and one for  $\gamma_1=[0.23]$ (training set). The results show that \ac{tpeqln} is able to correctly predict the trajectory for unobserved obstacle heights while keeping the desired shape and reaching the desired highest point of the curve. On the other hand, \ac{cnmp} accurately reproduces the trajectories from the training data set but generates larger distortions in the extrapolation domain, especially for the highest point of the trajectory.
This can be seen in Fig.~\ref{fig:results_OB_1D}~(middle), where the MSE for each feature value of each method is visualized. Here, it is observed that our method is the one that presents the smallest MSE for each value of $\gamma_1$. Finally, Fig.~\ref{fig:results_OB_1D}~(bottom) presents the MSE of the highest point of the curve. This point is of  importance to avoid the collision with the obstacle. The plots show that \ac{tpeqln} outperforms both \ac{tpgmm} and \ac{cnmp} in both training and extrapolation sets. 


It is worth mentioning that the two frames used by the \ac{tpgmm} at the beginning/end of the motion are constant for each value of the feature parameter, while the frame in the middle varies depending on the height of the obstacle. Otherwise the estimated trajectory from  \ac{tpgmm} will be constant along the $z$-axis. In other words, for \ac{tpgmm} we assume that the highest point to reach is always known, even for the extrapolation domain. In practice, this point is unknown and has to be estimated since it is directly correlated with the height of the obstacle. 
For the \ac{tpeqln} and \ac{cnmp} methods, this extra information is not provided. This assumption reveals a first clear advantage of our method over \ac{tpgmm}, i.e., it can be more practical to specify a set of high-level feature parameters rather than a reference frame.


\subsection{$\Phi_2$ - Variable height and goal}

\begin{figure}[t]
    \centering
    \includegraphics[width=\columnwidth]{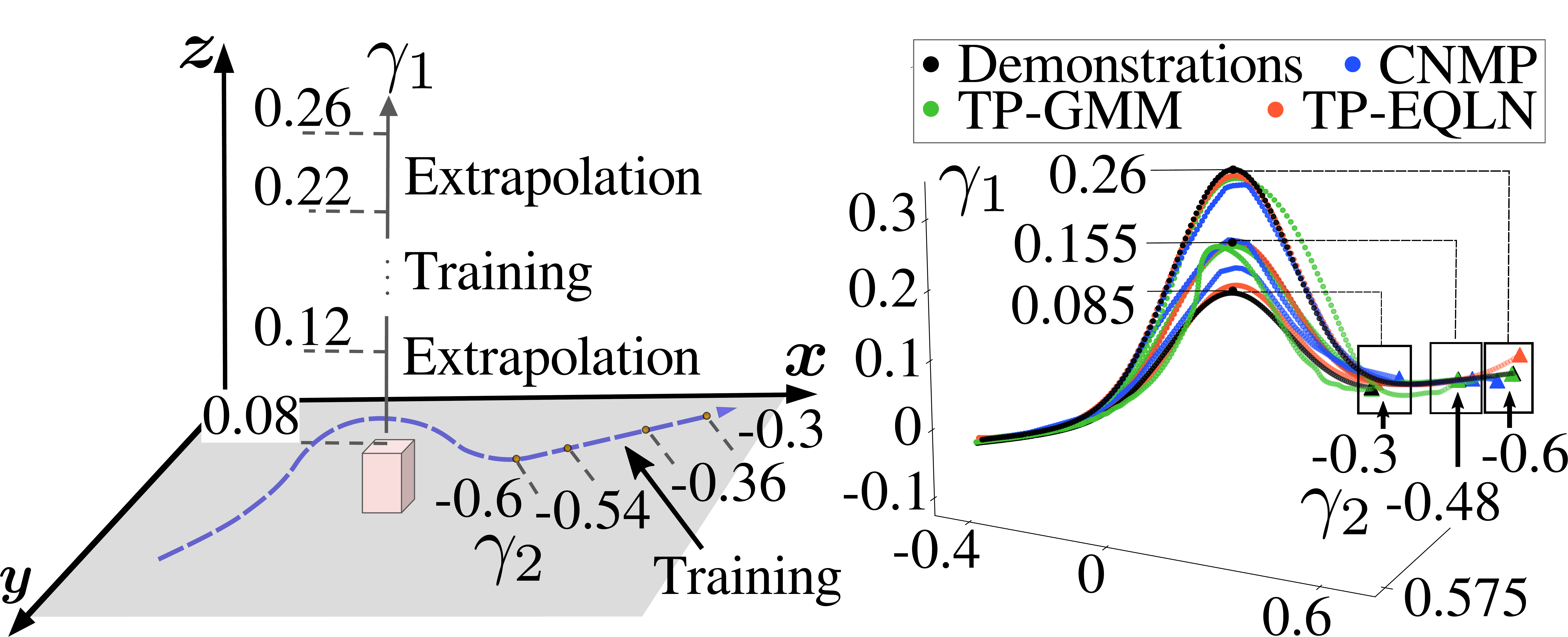}

    \vspace{4mm}
    
    \includegraphics[width=1.0\columnwidth]{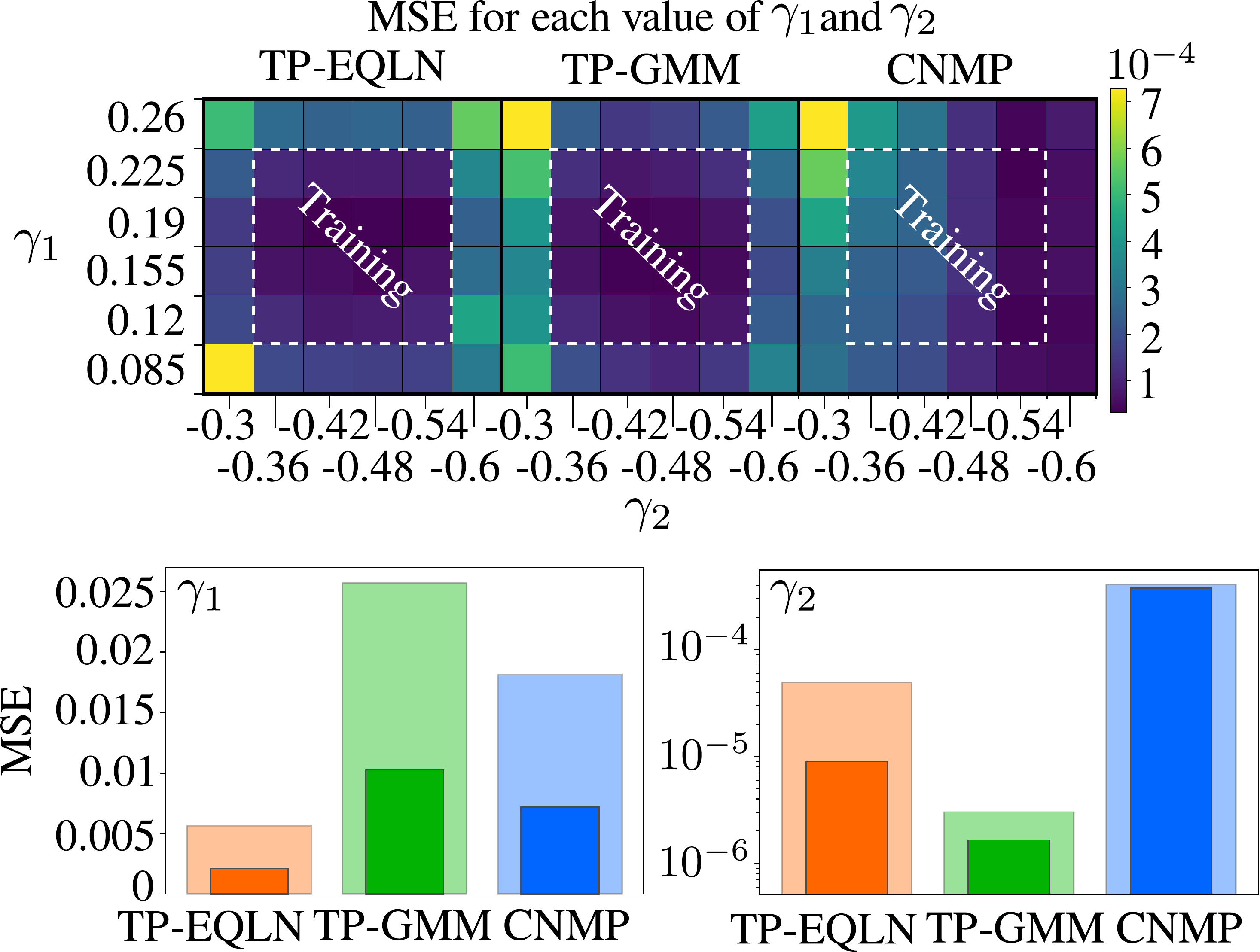}
    \caption{Definition of the task $\Phi_2$ and qualitative results. (Top-left) The feature space considered for the task. (Top-right) Three trajectories, one from the training set and two for the validation set, obtained for each method. (Bottom) MSE estimation for each approach for both reproduction and validation set.}
    \label{fig:Definition_and_trajectory_OB_2D}
\end{figure}

Similarly to the previous task, we vary the height of the obstacle encoded in $\gamma_1$. Additionally, we introduce a new task parameter $\gamma_2$ to encode the final position (goal). The feature parameter vector is formed as $\vec{\gamma}=[\gamma_1,\gamma_2]$. 
The demonstrations were sampled from~\eqref{eq:eq_update_exp}, with the parameters defined in Table~\ref{table:Feature_parameters_tasks}. 
%
An overview of this task is depicted in Fig.~\ref{fig:Definition_and_trajectory_OB_2D}~(top-left), while $3$ trajectories retrieved with each approach are shown in Fig.~\ref{fig:Definition_and_trajectory_OB_2D}~(top-right). Inb particular, we show two trajectories from the extrapolation set with $[\gamma_{1},\gamma_{2}]=[[0.085,-0.3],[0.26,-0.6]]$ and one from the training  set with $[\gamma_{1},\gamma_{2}]=[0.155,-0.48]$. For these cases, the results show that our approach presents less distortion in the shape of the trajectory than the other  methods (Fig.~\ref{fig:Definition_and_trajectory_OB_2D}, bottom).

\subsection{$\Phi_3$ - Variable height, obstacle position, and goal}
\begin{figure}[t]
    \centering
    \includegraphics[width=\columnwidth]{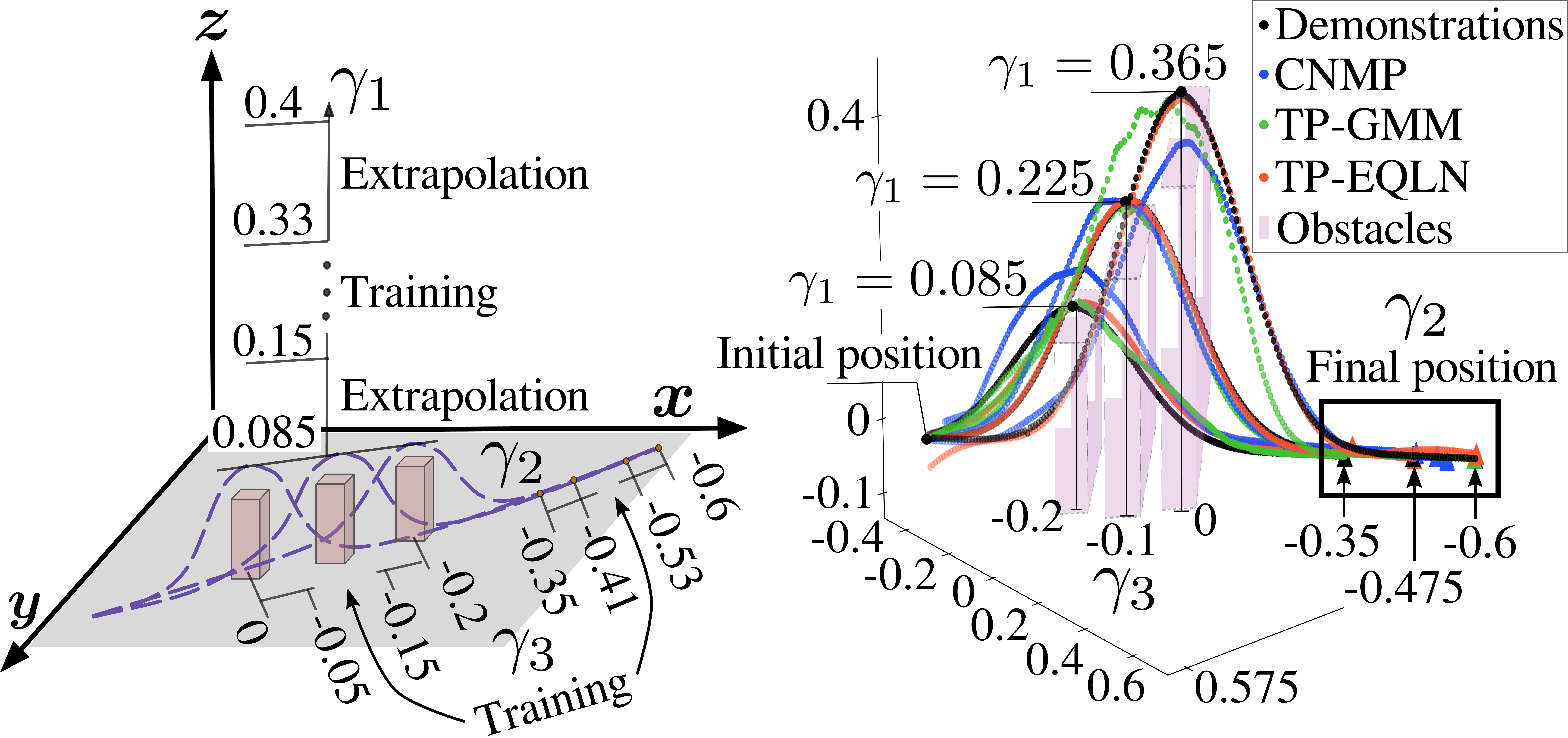}

    \vspace{2mm}
    
    \includegraphics[width=1.0\columnwidth]{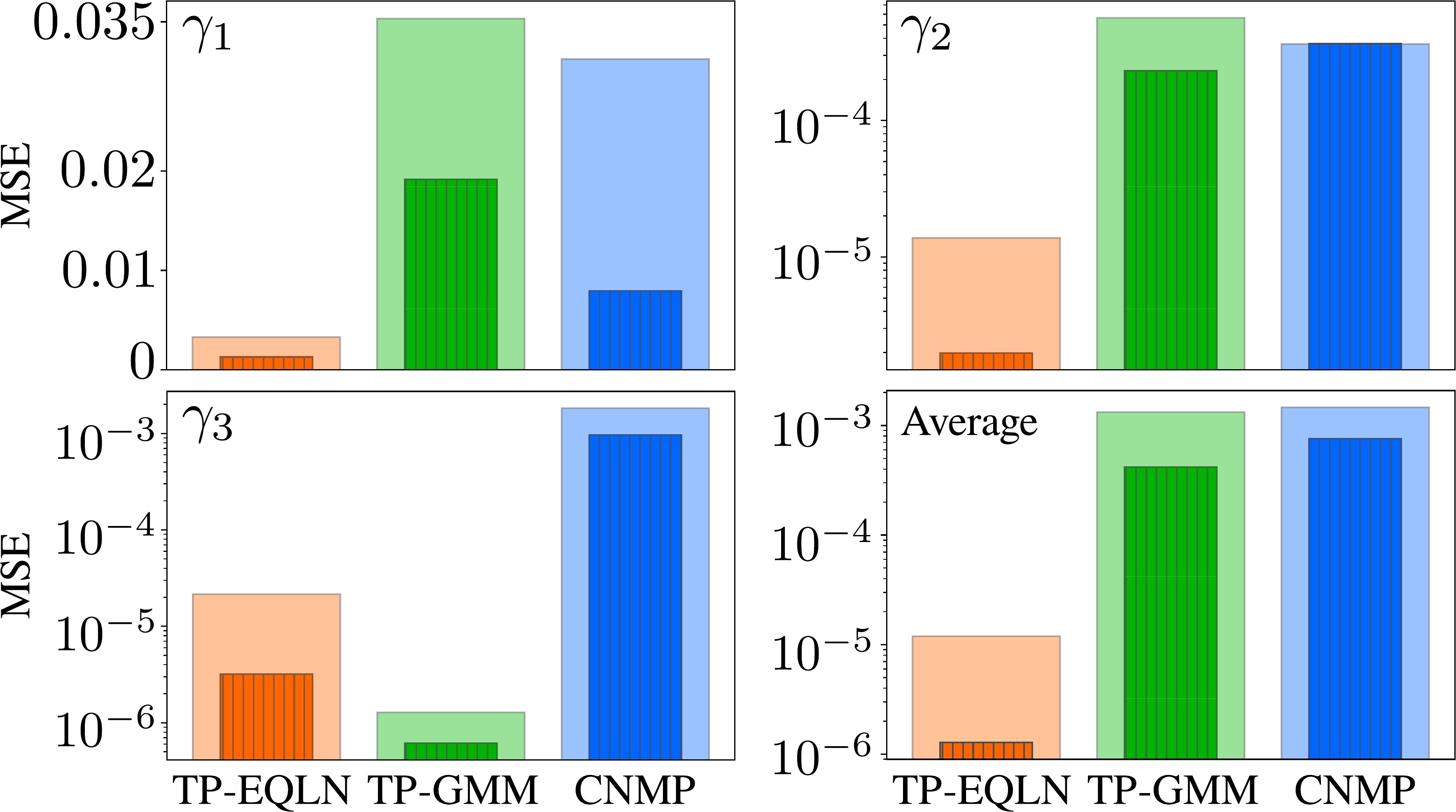}
    \caption{Definition of the task $\Phi_3$ and obtained results. (Top-left) The feature space considered for the task. (Top-right) Three trajectories, one from the training set and two for the validation set, obtained for each method. (Bottom) MSE obtaned for each method considering all the demonstrations in the $\Phi_3$ task. Shaded (non-shaded) areas indicate the extrapolation (training) domain.}
    \label{fig:description_and_results_OB_3D}
\end{figure}

In this task we change the obstacle height, its position, and the final point of the trajectory, which are embedded in three feature parameters $\vec{\gamma}=\left [\gamma_1,\gamma_2,\gamma_3\right ]$. Demonstrations are generated from~\eqref{eq:eq_update_exp}, with the parameters defined in Table~\ref{table:Feature_parameters_tasks}. 
The position of the obstacle is also varied linearly by shifting the Gaussian function with $\mu$, where $\mu$ depends on $\gamma_3$. The results of this procedure are shown in Fig.~\ref{fig:description_and_results_OB_3D}~(top-left). 

Fig.~\ref{fig:description_and_results_OB_3D}~(top-right) shows $3$ predicted trajectories for each method with three different sets of feature parameter values, i.e. $\vec{\gamma}=[0.225,-0.475,-0.1]$ (training set) and $\vec{\gamma}=[[0.085,-0.35,-0.2],[0.365,-0.6,0.0]]$ (extrapolation set).
For \ac{tpgmm}, we have provided again $3$ reference frames. The frame at the beginning remains constant, while the others vary depending on the desired obstacle's position and the goal of the trajectory. As done in $\Phi_1$ and $\Phi_2$, the position of the frames is provided also for parameter values in the extrapolation domain. 

Figure~\ref{fig:description_and_results_OB_3D}~(bottom) shows the MSE for $\vec{\gamma}$. The MSE for $\gamma_1$ represents error of the highest point, the MSE for $\gamma_2$ represents error of the obstacle's position, which is measured with respect to the position of the highest point of the trajectory in the plane $xz$, and the MSE for $\gamma_3$ represents error of the final goal position. These results show that our approach reaches closer to the highest point of the trajectory than the other methods. Regarding the MSE for $\gamma_2$, our method is slightly outperformed by \ac{tpgmm}. The reason can be the extra information provided by the middle frame. Regarding the MSE for $\gamma_3$ the \ac{tpeqln} outperforms other methods. This means that our method is able to predict better the desired highest point of the trajectory at the right place to prevent collisions. 

%
Finally, the average MSE over all the trajectories (for both training and extrapolation set) show that \ac{tpeqln} is able to preserve the shape of the trajectory even for feature parameter values in the extrapolation domain. 

The presented results show that our method predicts the spatial features linked to the features parameters accurately, while maintaining the shape of the trajectory even for feature parameters values from the extrapolation domain.

\begin{figure}[t]
    \centering
    \includegraphics[width=\columnwidth]{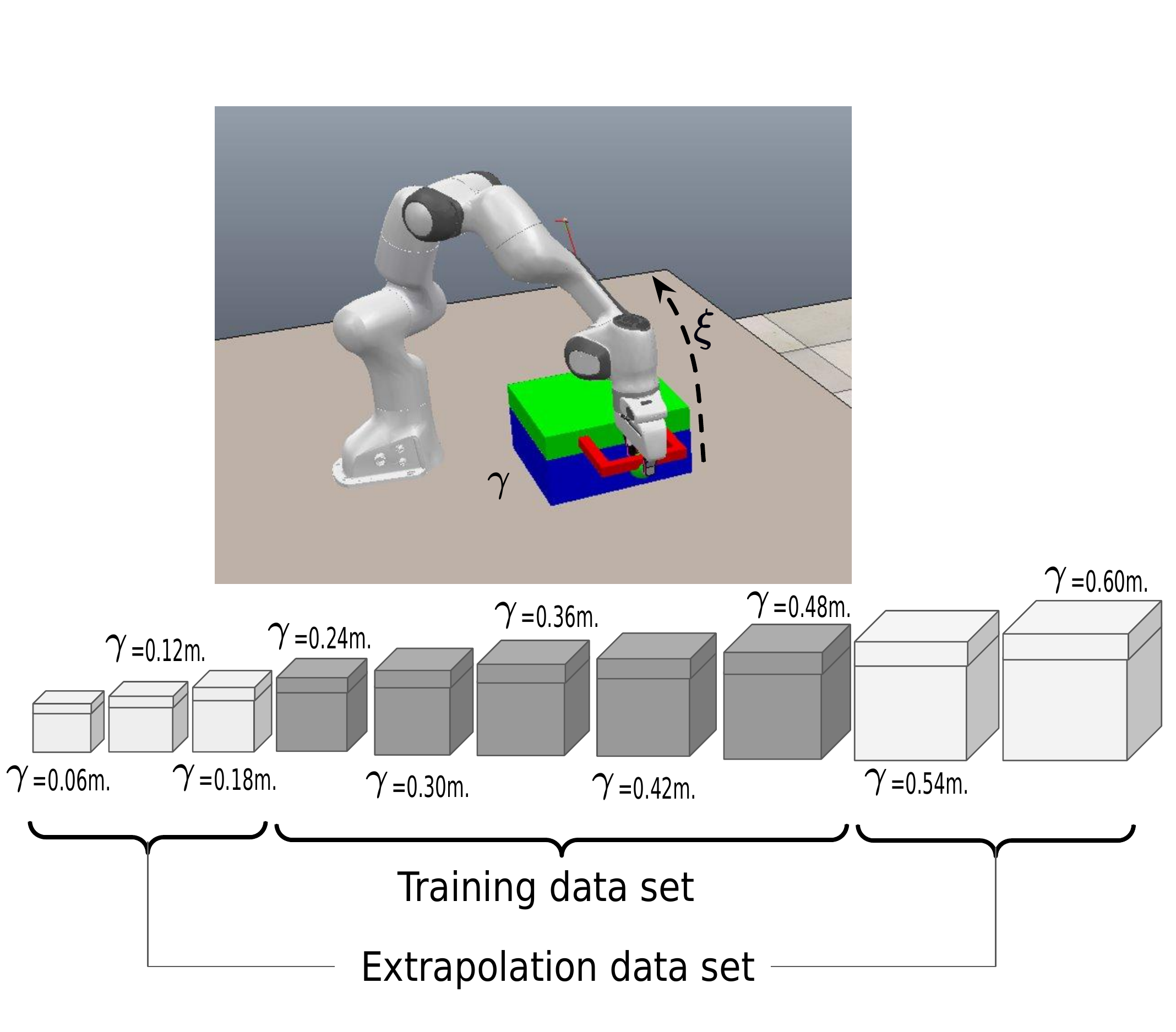}
    
    \vspace{2mm}
    
     \includegraphics[width=\columnwidth]{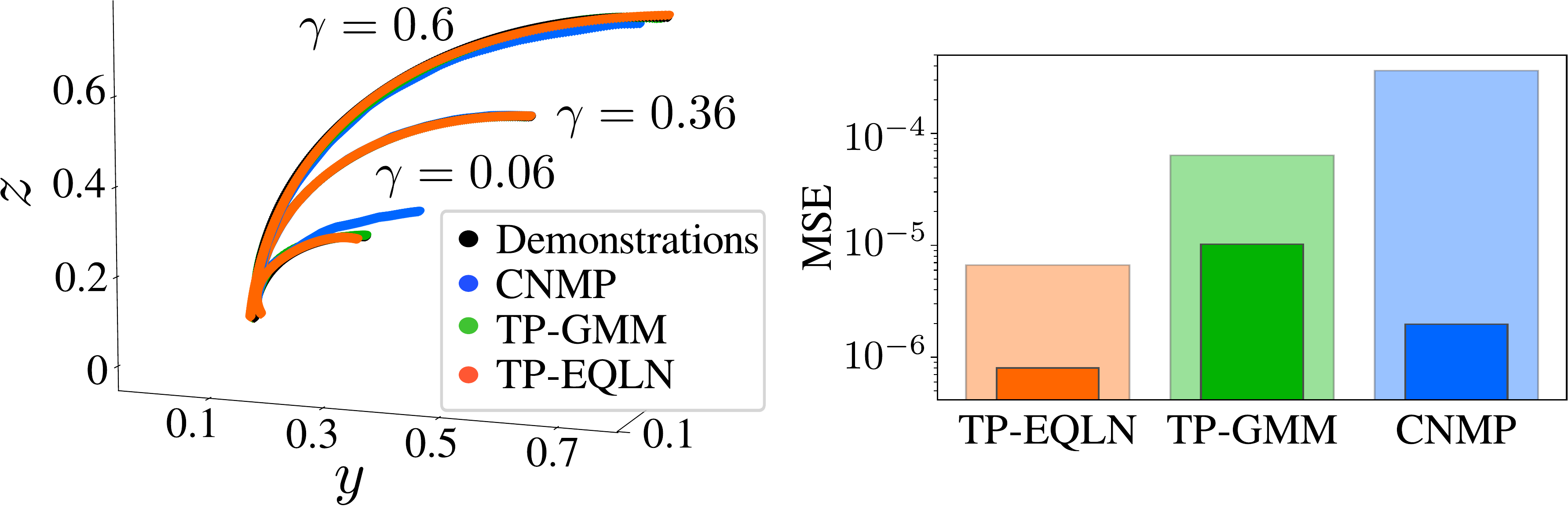}
    \caption{Definition of Task $\Phi_4$. (Top) Each feature parameter encodes the trajectory to be performed depending on the box' size. (Bottom-left) Three  Predicted trajectories one from the training set and two for extrapolation set. (Bottom-right) MSE obtained for each method. Shaded (non-shaded) areas indicate the extrapolation (training) domain.}
    \label{fig:Open_box_task}
\end{figure}
\subsection{$\Phi_4$ - Opening a box}
This task consists in opening boxes of different sizes by holding their cover and executing an arch trajectory until the box is entirely open (Fig.~\ref{fig:Open_box_task}, top). The box's size is given as a feature parameter $\gamma$, which affects both the trajectory's arc that must be followed as well as its endpoint.
 The trajectories are expressed with respect to the center of the handle and are generated as:
\begin{equation}\label{eq:open_box_motion}
\begin{split}
&x=0\\
&y=(0.15+\gamma)\left(1-\cos\left(\frac{t\pi}{400}\right)\right)\\
&z=(0.15+\gamma)\sin\left(\frac{t\pi}{400}\right)
\end{split},
\end{equation}
where the value of the parameters is defined in  Table~\ref{table:Feature_parameters_tasks}.
We provide two frames for \ac{tpgmm},  one at the beginning and one at the end of the trajectory. The frame at the beginning remains constant, whereas the frame at the end varies depending on the feature parameter. The frame value is considered as extra information since in practice the end point of the trajectory is unknown in the extrapolation domain.
Fig.~\ref{fig:Open_box_task}~(bottom-left) shows the predicted trajectory for $\gamma=0.06,0.6$ (extrapolation set) and for $\gamma=0.36$ (training set). In the presented cases,  \ac{cnmp} is the one that shows larger deviations from the desired trajectories in extrapolation whereas \ac{tpeqln} and \ac{tpgmm} predict accurately the shape of the trajectory. This can be seen in Fig.~\ref{fig:Open_box_task}~(bottom-right), where the MSE for both training and extrapolation set are shown. 

    
    
%
\subsection{$\Phi_5$ - Pick and Place with Obstacle avoidance}
\begin{figure*}[t]
  \includegraphics[width=1.0\textwidth]{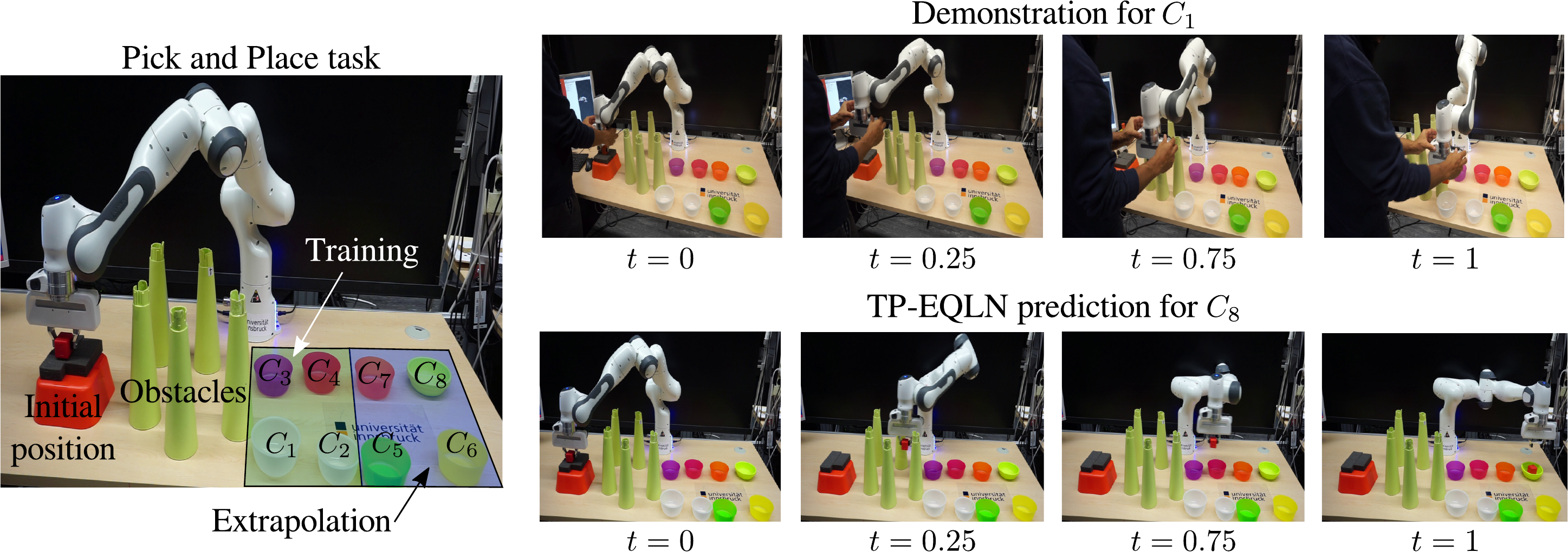}
  
  \vspace{2mm}
  
  \includegraphics[width=1.0\textwidth]{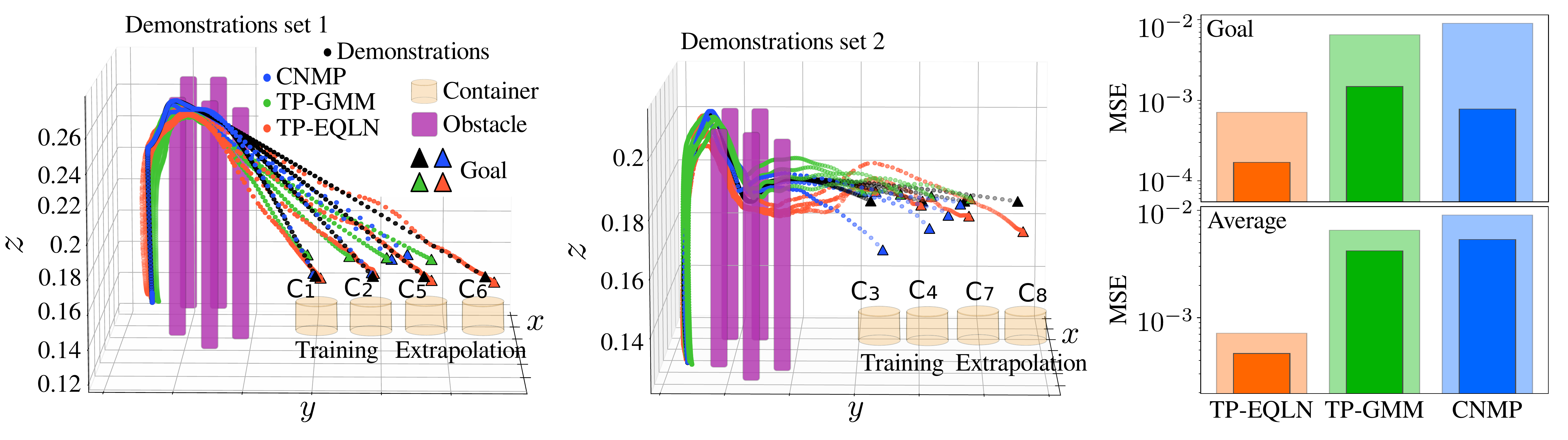}
  \caption{Definition of task $\Phi_5$ and qualitative results. (Top-left) The experiment setup. (Bottom-left) Trajectories obtained from each method
  in the training (containers $C_1$,$C_2$) and extrapolation data sets (containers $C_5$, $C_6$). (Bottom-center) Trajectories obtained from each method
  in the training (containers $C_3$,$C_4$) and extrapolation data sets (containers $C_7$, $C_8$).} (Bottom-right) MSE estimation for each approach. Shaded (non-shaded) areas indicate the extrapolation (training) domain.
   \label{fig:Pick_Place_exp_snapshots}
\end{figure*}

This experiment evaluates the prediction capabilities of our approach in a real pick and place task. The experiment consists of releasing an object in a container while avoiding some fixed obstacles in the environment (see Fig.~\ref{fig:Pick_Place_exp_snapshots}, top-left). We used 8 containers for this experiment denoted as $C_i$ with $i\in\left \{1,\ldots,8  \right \}$. However, we demonstrate the pick and place task only for the containers $i\in\left \{1,2,3,4 \right \}$. The initial position of the object to be transported is fixed whereas the release point is defined right above of each container's position. The demonstrations were collected using a Panda robot, where the user demonstrated kinesthetically the trajectories, as shown in Fig.~\ref{fig:Pick_Place_exp_snapshots}, top-right).
Each trajectory was stored with a resolution of $700$ data-points and smoothed using Cubic Splines from the SciPy library. The preservation of the trajectories' shape in this task is  relevant to guarantee the avoidance of the obstacles in the setup.

The feature parameter is defined with the initial and the release point, i.e.\ $\gamma_i \in \mathbb{R}^{6}$. Although the initial point remains constant, we found out that \ac{tpeqln} performs better by including this initial point in the feature parameter. The feature parameter space and the training and extrapolation data sets considered for this task can be found in Table~\ref{table:Feature_parameters_tasks}. 

For \ac{tpgmm}, we have defined two frames, at the beginning and at the end of each trajectory.
In Fig.~\ref{fig:Pick_Place_exp_snapshots}~(bottom-left and middle), we show a trajectory in training and $2$ in extrapolation domain executed by the robot with \ac{tpgmm}, \ac{cnmp}, and \ac{tpeqln}. The $3$ methods produce correct trajectories for avoiding the obstacles successfully in both data-sets. For the release point, the three methods present a gap with respect to the goal. This gap is mainly present in the $z$ axis.  

%

\begin{figure*}[t]
  \includegraphics[width=\textwidth]{Figures/Pouring_experiment_compressed.pdf}
  
  \vspace{2mm}

  \centering
  \includegraphics[width=0.7\textwidth]{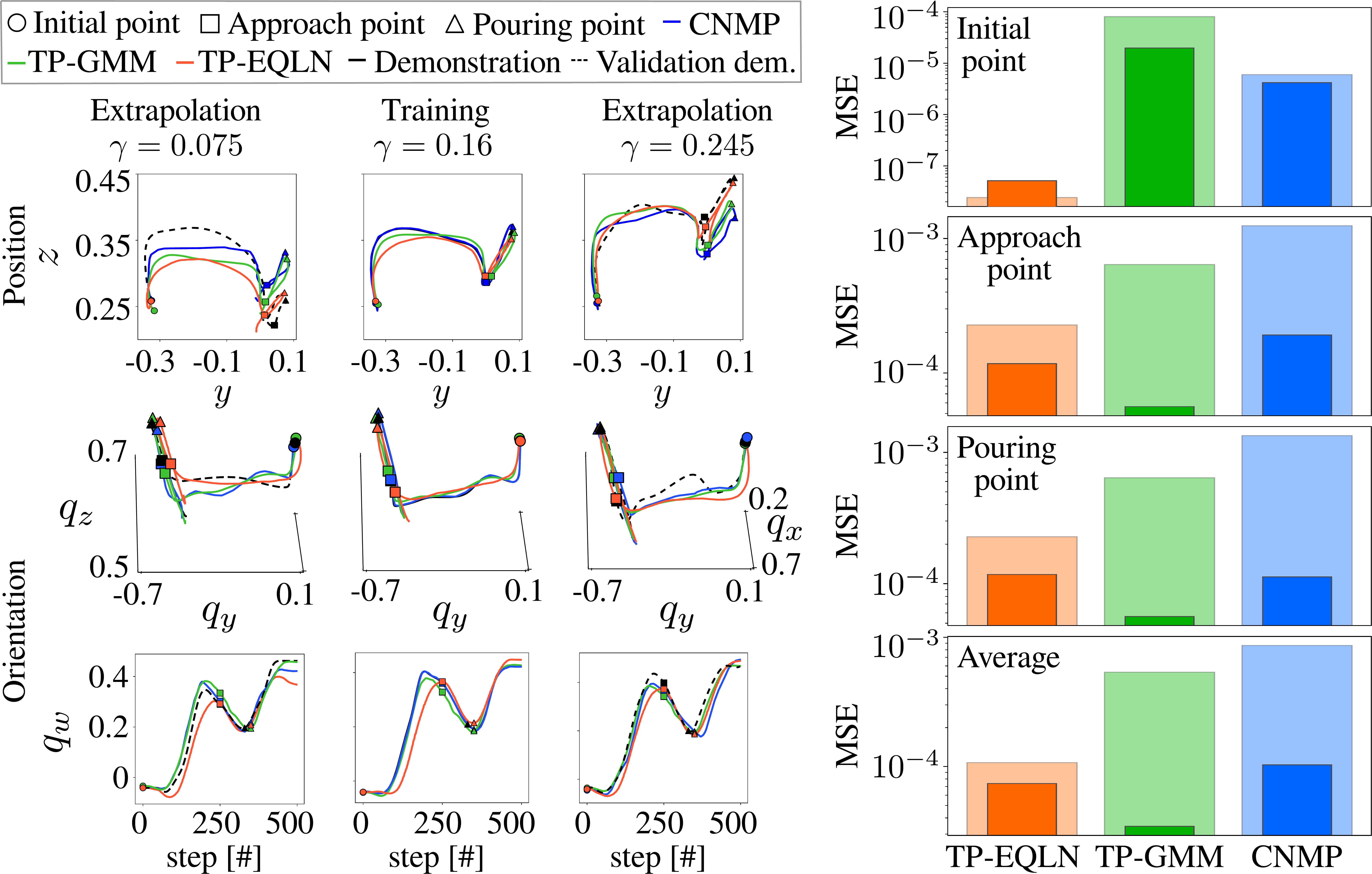}
  \caption{Definition of the task $\Phi_6$ and results. (Top-left) the experiment setup. 
  (Top-right) Snapshots of a task demonstration and reproduction (in extrapolation) using \ac{tpeqln}. (Bottom-left) Three trajectories for position and orientation obtained for each method, where we highlight the most important (key) points of each demonstration.  
  (Bottom-right) MSE obtained for each key points and considering the whole trajectory for training/extrapolation (non-shaded/shaded ares) sets.}
  \label{fig:Pouring_exp_snapshots}
\end{figure*}
Figure~\ref{fig:Pick_Place_exp_snapshots}~(bottom-right) shows the MSE of the release point (goal) for each data set. Results show that \ac{tpeqln} produces the lowest gap in both training and extrapolation sets. This means that \ac{tpeqln} releases the object in the container successfully, whereas the \ac{tpgmm} and \ac{cnmp} release the object outside of the container. 
The shape of the trajectory is also important for executing the task successfully.
In Fig.~\ref{fig:Pick_Place_exp_snapshots}~(bottom-right), we show the average MSE for each data set which tell us how well the shape of the trajectory is preserved. The results show that \ac{tpeqln} significantly reduces the total error in reproducing the demonstrations. 
Figure~\ref{fig:Pick_Place_exp_snapshots}~ (middle-right) shows some snapshots of the executed trajectory for the Container $C_8$ using \ac{tpeqln}.
\subsection{$\Phi_6$ - Pouring into different cups}
This experiment demonstrates the generalization capabilities of \ac{tpeqln} in a pouring task (see Fig.~\ref{fig:Pouring_exp_snapshots}, top-left), where both position and orientation of the end-effector have to be controlled. In particular, the position trajectory is mainly affected by the size of the cup and therefore varies in the $xy$ plane. The orientation plays an important role as it gives the correct inclination of the end-effector during the pouring, but it does not present a significant variation with respect to the size of the cup. Also the correct coupling between position and orientation trajectories is important for the successful execution of the task.  The estimated trajectory is $\xi\in \mathbb{R}^{7}$, where $[x,y,z]$ defines the position of the end-effector in Cartesian space and $[q_x,q_y,q_z,q_w]$ the orientation as a unit quaternion. The initial position of the cup is constant for all the demonstrations. The variation of the feature parameter $\gamma\in\mathbb{R}$ is linked to the cup's size and the ranges are listed in Table~\ref{table:Feature_parameters_tasks}. 

The demonstrations were taken with a Panda robot where the user demonstrated kinesthetically the trajectories as shown in the snapshots of Fig.~\ref{fig:Pick_Place_exp_snapshots} (top-right). Each trajectory was stored with a resolution of $700$ data points. We  estimate position and orientation using separate \acp{tpeqln} or \ac{cnmp} that share the feature parameter as input. Both estimated trajectories are aligned in time to form the estimated trajectory $\hat{\xi}\in \mathbb{R}^{7}$.

 
As done in the other task for comparison purposes, we provide to \ac{tpgmm} information about the unknown (final) point by specifying the value of the end frame also in extrapolation.

In Fig.~\ref{fig:Pouring_exp_snapshots}~(bottom-left), we show the predicted trajectories for position and orientation for $3$ different cup sizes, i.e.\ $\gamma=0.16$ (training set) and $\gamma=[0.075, 0.245]$ (extrapolation set). For this task, there are three relevant points of the trajectory (see the markers in Fig.~\ref{fig:Pouring_exp_snapshots}, bottom-left). These points are used as a reference to compare the \ac{tpeqln} against \ac{tpgmm} and \ac{cnmp}. The first point is the initial pose of the robot. All the approaches manage to keep the initial pose for different feature parameters. The second point represents the approach position between the tip of the watering can and the cup. The last point defines the pose where the robot ends the pouring process. The last one is the most important considering that a deviation in this point may cause the liquid to spill outside the cup.

In Fig.~\ref{fig:Pouring_exp_snapshots}~(bottom-right), we show the MSE value for each key point in both training and extrapolation domains. Regarding the initial point, our method is the one that presents the lowest MSE. A large MSE in the initial point of the trajectory would generate large control actions and a non-smooth start of the robot motion. Regarding the approach and final point, our method is slightly outperformed by  \ac{tpgmm} in the training domain. However, in the extrapolation domain, our method produces better results. In Fig.~\ref{fig:Pouring_exp_snapshots}~(bottom-right), we show the average MSE per data set. In the training domain, \ac{tpeqln} and \ac{cnmp} perform similarly and are slightly outperformed by \ac{tpgmm}. \ac{tpeqln} outperforms other approaches in extrapolation.

\section{Conclusions and Future Work}\label{sec:conclusion}
We presented the \acf{tpeqln}, a novel approach to imitation learning with enhanced extrapolation capabilities for tasks where maintaining the shape of the trajectory is important. The key idea of  \ac{tpeqln} is to combine different types of activation functions in order to find an analytical expression that fits the training data. This leads to a better approximation of the desired motion in both inter- and extrapolation domains. We enrich our network with task-dependent parameters and use them alongside a time vector to query the desired path. 
We showed it is possible to encode physical or spatial properties of the task in this feature parameter and to generalize the task over the feature parameter space.

Our \ac{tpeqln} was empirically evaluated and compared against two prominent existing approaches in a set of tasks of increasing complexity. The obtained results show that, in most of the cases, \ac{tpeqln} outperforms state-of-the-art approaches, especially in extrapolation, 
and most importantly, preserves the shape of the trajectory. Whereas \ac{cnmp} also use a feature parameter to encode some properties of the task, it is not possible to generalize beyond the data distribution of the training data since this method is based on probability distributions. On the other hand, \ac{tpgmm} require the explicit definition of frames provided by the user to be able to adapt the trajectory for any change in the environment. \ac{tpeqln} encode those changes in the environment using high-level feature parameters. With this is possible to generalize the task beyond the data distribution of the feature parameter used for training. The definition of high-level feature parameters
allows us to define and generalize the tasks more intuitively.

Our future research will follow different directions. We will explore the possibility to combine  \ac{tpeqln} with control techniques to ensure convergence to a given target while keeping the overall shape of the motion. We also plan to endow \ac{tpeqln} with stochastic information to characterize the variability and the uncertainty in the demonstrations. Finally, we will investigate the possibility to use more abstract concepts as task parameters, like the property of containing liquids, in combination with a task ontology to generalize learned skills across different objects with similar properties.

\section*{Acknowledgements}
This research has received funding from the Austrian Research Foundation (Euregio IPN 86-N30, OLIVER).

The authors thank M. Yunus Seker and E. Ugur for sharing the CNMP code, which is now publicly available \cite{CNMP_Code}.

\bibliographystyle{elsarticle-num}
\bibliography{bibliography.bib}

\begin{thebibliography}{10}
\expandafter\ifx\csname url\endcsname\relax
  \def\url#1{\texttt{#1}}\fi
\expandafter\ifx\csname urlprefix\endcsname\relax\def\urlprefix{URL }\fi
\expandafter\ifx\csname href\endcsname\relax
  \def\href#1#2{#2} \def\path#1{#1}\fi

\bibitem{billard2016learning}
A.~Billard, S.~Calinon, R.~Dillmann, Learning from humans, in: B.~Siciliano,
  O.~Khatib (Eds.), Handbook of Robotics, Springer, Secaucus, NJ, USA, 2016,
  Ch.~74, pp. 1995--2014, 2nd Edition.

\bibitem{saveriano2015incremental}
M.~{Saveriano}, S.~{An}, D.~{Lee}, Incremental kinesthetic teaching of
  end-effector and null-space motion primitives, in: IEEE International
  Conference on Robotics and Automation, 2015, pp. 3570--3575.

\bibitem{caccavale2019kinesthetic}
R.~Caccavale, M.~Saveriano, A.~Finzi, D.~Lee, Kinesthetic teaching and
  attentional supervision of structured tasks in human--robot interaction,
  Autonomous Robots 43~(6) (2019) 1291--1307.

\bibitem{ude2010task}
A.~{Ude}, A.~{Gams}, T.~{Asfour}, J.~{Morimoto}, Task-specific generalization
  of discrete and periodic dynamic movement primitives, IEEE Transactions on
  Robotics 26~(5) (2010) 800--815.

\bibitem{calinon2016tutorial}
S.~Calinon, A tutorial on task-parameterized movement learning and retrieval,
  Intelligent Service Robotics 9~(1) (2016) 1--29.

\bibitem{pervez2018learning}
A.~Pervez, D.~Lee, Learning task-parameterized dynamic movement primitives
  using mixture of gmms, Intelligent Service Robotics 11~(1) (2018) 61--78.

\bibitem{MartiusL16}
G.~Martius, C.~H. Lampert, \href{http://arxiv.org/abs/1610.02995}{Extrapolation
  and learning equations}, CoRR abs/1610.02995.
\newblock \href {http://arxiv.org/abs/1610.02995} {\path{arXiv:1610.02995}}.
\newline\urlprefix\url{http://arxiv.org/abs/1610.02995}

\bibitem{pmlr-v80-sahoo18a}
S.~Sahoo, C.~Lampert, G.~Martius, Learning equations for extrapolation and
  control, in: International Conference on Machine Learning, 2018, pp.
  4442--4450.

\bibitem{Calinon2007learning}
S.~{Calinon}, F.~{Guenter}, A.~{Billard}, On learning, representing, and
  generalizing a task in a humanoid robot, IEEE Transactions on Systems, Man,
  and Cybernetics, Part B (Cybernetics) 37~(2) (2007) 286--298.

\bibitem{Khansari2011learning}
S.~M. Khansari-Zadeh, A.~Billard, Learning stable non-linear dynamical systems
  with gaussian mixture models, Transactions on Robotics 27~(5) (2011)
  943--957.

\bibitem{Perrin16fast}
N.~Perrin, P.~Schlehuber-Caissier, Fast diffeomorphic matching to learn
  globally asymptotically stable nonlinear dynamical systems, Systems \&
  Control Letters 96 (2016) 51--59.

\bibitem{saveriano2018incremental}
M.~Saveriano, D.~Lee, Incremental skill learning of stable dynamical systems,
  in: IEEE/RSJ International Conference on Intelligent Robots and Systems,
  IEEE, 2018, pp. 6574--6581.

\bibitem{saveriano2020energy}
M.~Saveriano, An energy-based approach to ensure the stability of learned
  dynamical systems, in: IEEE International Conference on Robotics and
  Automation, IEEE, 2020, pp. 4407--4413.

\bibitem{Ijspeert2013Dynamical}
A.~J. Ijspeert, J.~Nakanishi, H.~Hoffmann, P.~Pastor, S.~Schaal, {Dynamical
  Movement Primitives: Learning Attractor Models for Motor Behaviors}, Neural
  Computation 25~(2) (2013) 328--373.

\bibitem{Saveriano2021Dynamic}
M.~Saveriano, F.~J. Abu{-}Dakka, A.~Kramberger, L.~Peternel, Dynamic movement
  primitives in robotics: {A} tutorial survey, CoRR abs/2102.03861.

\bibitem{Rasmussen2006}
C.~E. Rasmussen, C.~K.~I. Williams, {Gaussian Processes for Machine Learning},
  The MIT Press, Cambridge, Massachusetts, 2006.

\bibitem{Stulp2013learning}
F.~Stulp, G.~Raiola, A.~Hoarau, S.~Ivaldi, O.~Sigaud, Learning compact
  parameterized skills with a single regression, in: IEEE-RAS International
  Conference on Humanoid Robots, Atlanta, GA, USA, 2013, pp. 417--422.

\bibitem{cohn1996active}
D.~A. Cohn, Z.~Ghahramani, M.~I. Jordan, Active learning with statistical
  models, Journal of artificial intelligence research 4 (1996) 129--145.

\bibitem{paraschos2013probabilistic}
A.~Paraschos, C.~Daniel, J.~Peters, G.~Neumann, Probabilistic movement
  primitives, in: C.~Burges, L.~Bottou, M.~Welling, Z.~Ghahramani,
  K.~Weinberger (Eds.), Advances in Neural Information Processing Systems 26,
  Curran Associates, Inc., Lake Tahoe, Nevada, US, 2013, pp. 2616--2624.

\bibitem{huang2019kernelized}
Y.~Huang, L.~Rozo, J.~Silv{\'e}rio, D.~G. Caldwell, Kernelized movement
  primitives, The International Journal of Robotics Research 38~(7) (2019)
  833--852.

\bibitem{Huang2020Toward}
Y.~{Huang}, F.~J. {Abu-Dakka}, J.~{Silv\'{e}rio}, D.~G. {Caldwell}, Toward
  orientation learning and adaptation in cartesian space, IEEE Transactions on
  Robotics 37~(1) (2021) 82--98.

\bibitem{Zhou2019Learning}
Y.~Zhou, J.~Gao, T.~Asfour, Learning via-point movement primitives with inter-
  and extrapolation capabilities, in: IEEE/RSJ International Conference on
  Intelligent Robots and Systems, Macau, China, 2019, pp. 4301--4308.

\bibitem{abu2015adaptation}
F.~J. Abu-Dakka, B.~Nemec, J.~A. J{\o}rgensen, T.~R. Savarimuthu,
  N.~Kr{\"u}ger, A.~Ude, Adaptation of manipulation skills in physical contact
  with the environment to reference force profiles, Autonomous Robots 39~(2)
  (2015) 199--217.

\bibitem{Yang2018dmp}
C.~Yang, C.~Zeng, C.~Fang, W.~He, Z.~Li, A dmps-based framework for robot
  learning and generalization of humanlike variable impedance skills, IEEE/ASME
  Transactions on Mechatronics 23~(3) (2018) 1193--1203.

\bibitem{abu2020variable}
F.~J. Abu-Dakka, M.~Saveriano, Variable impedance control and learning—a
  review, Frontiers in Robotics and AI 7 (2020) 590681.

\bibitem{Rozo2016Learning}
L.~Rozo, S.~Calinon, D.~G. Caldwell, P.~Jiménez, C.~Torras, Learning physical
  collaborative robot behaviors from human demonstrations, IEEE Transactions on
  Robotics 32~(3) (2016) 513--527.

\bibitem{kronander2013learning}
K.~Kronander, A.~Billard, Learning compliant manipulation through kinesthetic
  and tactile human-robot interaction, IEEE transactions on haptics 7~(3)
  (2013) 367--380.

\bibitem{jaquier2021tensor}
N.~Jaquier, R.~Haschke, S.~Calinon, Tensor-variate mixture of experts for
  proportional myographic control of a robotic hand, Robotics and Autonomous
  Systems 142 (2021) 103812.

\bibitem{fang2022semg}
B.~Fang, C.~Wang, F.~Sun, Z.~Chen, J.~Shan, H.~Liu, W.~Ding, W.~Liang,
  Simultaneous semg recognition of gestures and force levels for interaction
  with prosthetic hand, IEEE Transactions on Neural Systems and Rehabilitation
  Engineering 30 (2022) 2426--2436.

\bibitem{peternel2018robot}
L.~Peternel, N.~Tsagarakis, D.~Caldwell, A.~Ajoudani, Robot adaptation to human
  physical fatigue in human--robot co-manipulation, Autonomous Robots 42~(5)
  (2018) 1011--1021.

\bibitem{Zeng2021Simultaneously}
C.~Zeng, C.~Yang, H.~Cheng, Y.~Li, S.-L. Dai, Simultaneously encoding movement
  and semg-based stiffness for robotic skill learning, IEEE Transactions on
  Industrial Informatics 17~(2) (2021) 1244--1252.

\bibitem{garnelo2018conditional}
M.~Garnelo, D.~Rosenbaum, C.~Maddison, T.~Ramalho, D.~Saxton, M.~Shanahan,
  Y.~W. Teh, D.~Rezende, S.~A. Eslami, Conditional neural processes, in:
  International Conference on Machine Learning, 2018, pp. 1704--1713.

\bibitem{seker2019conditional}
M.~Y. Seker, M.~Imre, J.~H. Piater, E.~Ugur, Conditional neural movement
  primitives, in: Robotics: Science and Systems, 2019.

\bibitem{akbulut2020acnmp}
M.~T. Akbulut, E.~Oztop, M.~Y. Seker, H.~Xue, A.~E. Tekden, E.~Ugur, Acnmp:
  Skill transfer and task extrapolation through learning from demonstration and
  reinforcement learning via representation sharing, in: Conference on Robot
  Learning, 2020.

\bibitem{schmidt2009distilling}
M.~Schmidt, H.~Lipson, Distilling free-form natural laws from experimental
  data, science 324~(5923) (2009) 81--85.

\bibitem{CNMP_Code}
M.~Yunus~Seker, \href{https://github.com/myunusseker/CNMP}{Conditional neural
  movement primitives ({CNMP}) - source code}.
\newline\urlprefix\url{https://github.com/myunusseker/CNMP}

\end{thebibliography}

\end{document}